\documentclass[preprint,authoryear,10pt,a4paper]{elsarticle}
\usepackage[top=0.75in, bottom=0.75in, left=0.75in, right=0.75in]{geometry}
\usepackage{palatino}
\usepackage{amsmath,amsthm}
\usepackage{amssymb}
\usepackage{bbm}
\usepackage{amsfonts}
\usepackage{graphicx}
\usepackage{caption}
\usepackage{subcaption}
\usepackage{grffile}
\graphicspath{{graphics/}}
\usepackage[ruled,vlined,linesnumbered]{algorithm2e}
\usepackage{tabularx,booktabs}
\usepackage{inputenc} 
\usepackage[T1]{fontenc}   
\usepackage{url}            
\usepackage{booktabs}       %
\usepackage{amsmath,amsthm}
\usepackage{amssymb}
\usepackage{bbm}
\usepackage{amsfonts}       
\usepackage{nicefrac}       
\usepackage{microtype}      
\usepackage{tikz}
\usepackage{rotfloat}
\usepackage{graphicx}
\usepackage{caption}
\usepackage[ruled,vlined]{algorithm2e}
\usepackage{algpseudocode}
\usepackage{setspace}
\usepackage{appendix}
\usepackage{tablefootnote}
\usepackage{bm}
\usepackage{mathrsfs}
\usepackage{xcolor}
\usepackage{colortbl}

\usepackage{makecell}

\usepackage{verbatim}
\usepackage{xcolor}
\definecolor{darkblue}{rgb}{0.0,0.5,0.5}
\definecolor{blue}{rgb}{0.0,0.59,0.84}
\definecolor{myblue}{RGB}{0,0,255}
  
\usepackage[colorlinks]{hyperref}
\hypersetup{colorlinks,breaklinks,linkcolor=blue,urlcolor=blue,anchorcolor=blue,citecolor=blue}
\usepackage{booktabs,caption}
\usepackage{multirow,bigdelim}
\usepackage{lscape}
\usepackage{lineno}
\usepackage{microtype}
\usepackage{soul}
\usepackage{lineno}
\usepackage{fancyhdr}

\usepackage{amsmath,amsfonts,bm}

\def\eqref#1{equation~\ref{#1}}

\def\1{\bm{1}}

\DeclareMathAlphabet{\mathsfit}{\encodingdefault}{\sfdefault}{m}{sl}
\SetMathAlphabet{\mathsfit}{bold}{\encodingdefault}{\sfdefault}{bx}{n}

\begin{document}

\begin{frontmatter}

\title{{\fontfamily{lmss}\selectfont {From Attacks to Curricula: Learnability-Guided Adversarial Training for Safe Autonomous Driving}}}


\author[label1]{Yuewen Mei}
\author[label1,label]{Tong Nie}
\author[label1]{Jie Sun}
\author[label1]{Haotian Shi}
\author[label]{Wei Ma}
\author[label1]{Jian Sun\corref{cor1}}
\ead{sunjian@tongji.edu.cn}

\address[label1]{College of Transportation \& Key Laboratory  of  Road  and  Traffic  Engineering of Ministry of Education, Tongji University, Shanghai, 201804, China}
\address[label]{Department of Civil and Environmental Engineering, The Hong Kong Polytechnic University, Hong Kong SAR, China}

\cortext[cor1]{Corresponding authors.}

\begin{abstract}
Closed-loop adversarial training has emerged as a vital paradigm for enhancing the safety of autonomous driving policies by enabling them to learn from rare safety-critical scenarios. Standard training pipelines typically generate adversarial scenarios first, then sample them for policy optimization.
However, most existing frameworks remain attack-oriented.
Driven primarily by collision maximization, current generators often synthesize practically unsolvable extreme situations, thereby degrading the learning process. Furthermore, conventional heuristic and simplified sampling strategies ignore the continuously evolving capability of the driving policy, leading to sample inefficiency and delayed convergence.
To overcome these limitations, we propose AlignADV, a learnability-guided closed-loop adversarial training framework designed to convert adversarial scenarios into resolvable and capability-aligned curricula.
First, we reformulate adversarial scenario generation as a preference alignment problem and employ direct preference optimization to guide the generator toward critical yet resolvable scenarios. Second, we introduce the concept of behavioral fingerprint to extract the intrinsic characteristics of the evolving policy and construct a multi-modal capability prediction model that accurately evaluates policy performance without expensive simulations. By combining the resolvability-aligned scenario set with these capability predictions, we develop a dynamic curriculum sampling mechanism that prioritizes scenarios targeting the exact vulnerabilities of the current policy. 
Comprehensive experiments using the Waymo Open Motion Dataset demonstrate that AlignADV significantly improves both convergence efficiency and final performance, reducing training steps by up to 40.6\% compared to baseline methods, while reducing collision rate and improving route completion rate in both normal and adversarial traffic conditions.
These results highlight a shift from attack-oriented scenario generation to learnability-guided policy improvement, offering a principled direction for safer and more efficient autonomous driving training. Project page and video demonstration: \url{https://meiyuewen.github.io/AlignADV/}.
\end{abstract}


\end{frontmatter}


\section{Introduction}
Autonomous driving (AD) technologies have achieved significant breakthroughs in recent years. In particular, end-to-end driving models have demonstrated potential in navigating urban and highway traffic environments. However, ensuring the reliability and safety of AD in complex and extremely rare safety-critical scenarios remains a severe challenge. 
Since traffic accidents occur with extremely low probabilities in the real world, relying solely on naturalistic driving data to train and test AD policy is insufficient \citep{feng2021intelligent}. To overcome the issue of rarity, simulation-based generation of safety-critical scenarios, particularly adversarial scenarios, has become a standard practice \citep{ding2023survey, xu2025diffscene}. Building upon this foundation, adversarial training has emerged as a potential closed-loop learning approach for improving driving safety \citep{hanselmann2022king, zhang2023cat, stoler2025rcg}. 
This paradigm models the environment as an attacker that challenges the AD policy, which acts as a defender, by maximizing collision risks. Through this two-player game, the AD policy learns from safety-critical interactions induced in simulation, thereby improving the driving performance in long-tail scenarios. 
Generally, the efficacy of this adversarial training paradigm depends on two key factors: what scenarios to generate and how to use these scenarios to maximize learning efficiency and policy performance.

\begin{figure*}
    \centering
    \includegraphics[width=0.95\linewidth]{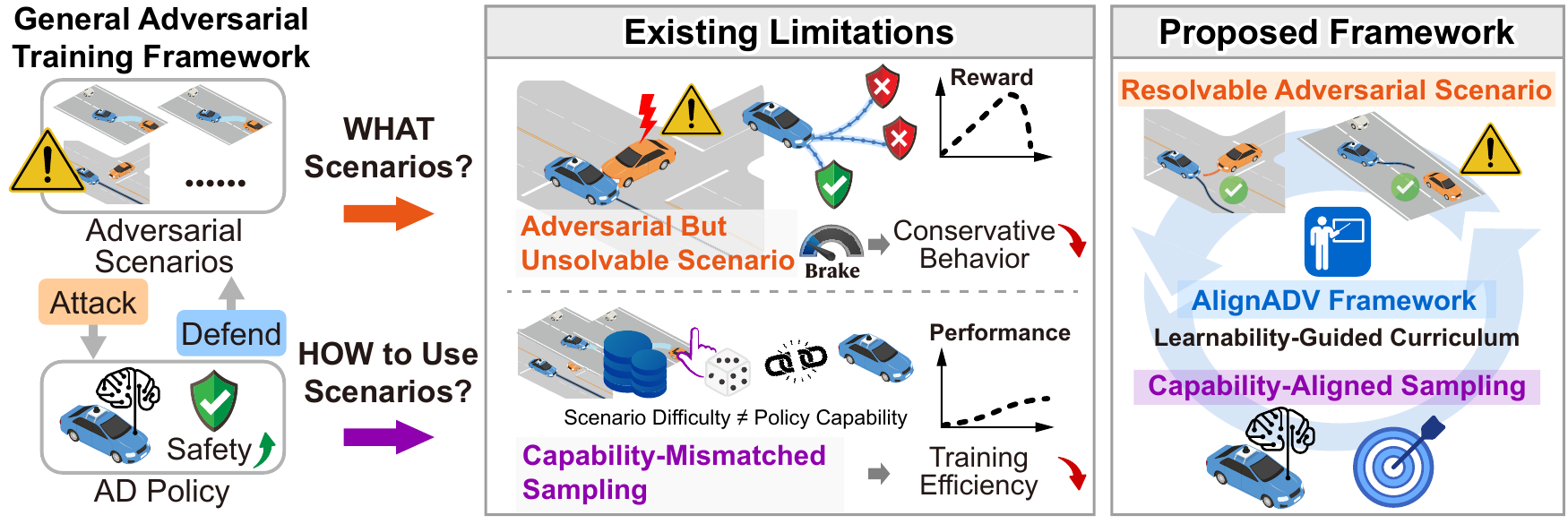}
    \caption{Overview of the research motivation and the proposed AlignADV framework. Existing methods exhibit two primary limitations: the generation of practically unsolvable scenarios that induce conservative driving behaviors and the capability-mismatched sampling that degrades training efficiency. Our proposed learnability-guided framework addresses these limitations by establishing a unified learning loop driven by resolvable adversarial scenarios and capability-aligned sampling strategies.}
    \label{fig:placeholder}
\end{figure*}


Regarding what scenarios to generate, existing studies have developed various generation methods, ranging from optimization-based methods \citep{hanselmann2022king, liu2025adv} to reinforcement learning \citep{liu2024safety, qiu2025aed} and deep generative models \citep{rempe2022generating, ding2023causalaf}. These methods commonly expose policy vulnerabilities by increasing collision risks, but this attack-oriented objective may also push generators toward overly aggressive scenarios.
Some recent methods introduce kinematic constraints and traffic rule regularizations to improve the physical realism and behavioral plausibility of generated scenarios \citep{chen2024frea, nie2025steerable}.
However, plausibility alone does not ensure resolvability: a trajectory may be rational for the background vehicle while leaving no reasonable collision-avoidance response for the ego vehicle.
Training on such unsolvable scenarios provides weak action-discriminative feedback and reward for policy learning, forcing the policy to learn overly conservative driving behaviors or even causing the collapse of the training process \citep{feng2026breaking}. 
Since traffic scenarios are high-dimensional and highly dynamic, it is difficult to explicitly characterize the differentiating boundary between critical yet resolvable challenges and unresolvable hazards. 
Therefore, the key challenge is how to guide adversarial generators toward safety-critical scenarios that remain practically resolvable and thus useful for policy learning.

In addition to the quality of the generated adversarial scenarios, how to effectively use these scenarios in the training loop is equally crucial. Existing training pipelines often employ static random sampling or rely on predefined scenario complexity metrics for heuristic sorting \citep{anzalone2022end, bronstein2023embedding, sheng2026curricuvlm}. Although some advanced methods consider automated curriculum selection, they mostly depend on feedback from historical interactions of the AD agent \citep{peng2024reward, abouelazm2025automatic}. 
However, as the parameters of the AD policy continuously update during training, its capability to handle different scenarios exhibits highly dynamic and non-linear fluctuations.
The aforementioned lagging sampling methods lack a forward-looking mechanism to evaluate the true competence of the current policy before actual interaction, leading to a severe mismatch between the difficulty of the sampled scenarios and the current capability of the policy. Such a mismatch weakens the learning process, as statically sampled scenarios may gradually become easy for the evolving policy.
This raises the key challenge of how to achieve a timely performance evaluation of AD policies during training over diverse scenario libraries without resorting to computationally expensive closed-loop simulations.


To overcome the aforementioned challenges, we propose a novel closed-loop adversarial training framework denoted as AlignADV. 
Instead of treating adversarial scenarios purely as attacks, AlignADV converts them into learnability-guided curricula. 
In this work, learnability serves as an operational principle that guides AlignADV to jointly pursue resolvability-aligned scenario generation and capability-aligned scenario sampling.
First, we reformulate the adversarial scenario generation as a preference alignment problem. We construct a preference dataset by introducing a rule-based expert policy and fine-tune the trajectory generation model via Direct Preference Optimization (DPO). As a post-training procedure, this alignment step directly reshapes the generator output distribution toward resolvable scenarios without modifying the specific generator architecture. 
Second, to make effective use of these resolvable scenarios during training, we propose a novel forward-looking policy capability prediction mechanism. We introduce the concept of the behavioral fingerprint for the AD policy to extract the intrinsic features of the continuously evolving policy network. By combining this fingerprint with the local context of specific scenarios, we design a cross-attention multi-modal prediction model to accurately output the predicted success probability of the current policy in any given scenario prior to actual simulation rollout.
Ultimately, based on these capability predictions, we construct a dynamic capability-aligned curriculum sampling distribution. 
This holistic framework ensures that the AD policy is challenged by solvable hazards that adaptively match its evolving capability, thereby avoiding both unsolvable scenarios and trivial tasks and enabling a more effective and efficient training curriculum for AD policies.
The main contributions of this paper can be summarized in the following three aspects:
\begin{itemize}
    \item We propose a resolvable adversarial scenario generation mechanism based on DPO, which serves as a general post-training fine-tuning paradigm for existing adversarial generation methods. It substantially improves the solvability of the generated adversarial environment while preserving high-risk challenges.
    \item We introduce the concept of a behavioral fingerprint and construct a multi-modal policy capability prediction model. The behavioral fingerprint provides an algorithm-agnostic representation of the driving policy. The capability prediction model efficiently evaluates the performance of the dynamically updating policy across diverse scenarios, offering a reliable quantitative metric for automated curriculum learning.
    \item We develop a closed-loop adversarial training framework, AlignADV, which couples resolvability-aligned generation with capability-aligned sampling to transform attack-oriented scenario generation into effective policy improvement. Compared to baseline method, this framework not only saves up to 40.6\% of the training steps but also significantly improves both collision rate and route completion rate of AD policy.
\end{itemize}

The remainder of this paper is organized as follows: Section \ref{sec:related_works} reviews the related works on adversarial training and curriculum learning; Section \ref{sec:Preliminary} formulates the problem for end-to-end driving and defines our adversarial training objective; Section \ref{sec:Methodology} elaborates on the methodology of the AlignADV framework, including resolvable scenario generation, capability prediction, and the curriculum sampling mechanism; Section \ref{sec:experiments} presents comprehensive experimental setups, result analyses, and ablation studies; and finally, Section \ref{sec:conclusion} concludes this work.








\section{Related Works}\label{sec:related_works}

\subsection{Adversarial Training}
Adversarial training has emerged as a prominent paradigm for enhancing the robustness of AD policies by exposing the driving agent to challenging and safety-critical scenarios. To overcome the rarity problem of safety-critical scenarios and efficiently supply these rare scenarios for policy training, numerous scenario generation methods have been developed. These methods can be broadly categorized into optimization-based, reinforcement learning (RL)-based, and generative model-based approaches.

Recently, optimization-based methods have been proposed to directly optimize the kinematic trajectories of background vehicles to collide with the ego vehicle \citep{hanselmann2022king, zhang2023cat, mei2025llm}. These methods are generally computationally efficient and maintain basic physical realism. Meanwhile, RL-based methods establish a Markov Decision Process to intelligently search for interactive adversarial strategies \citep{ransiek2024goose, cai2024vcat}. By dynamically maximizing the collision probability, RL approaches specialize in exploring a wide range of complex attack behaviors. Additionally, recent studies utilize deep generative models, such as diffusion models, to create hazardous scenarios by conditioning the generation on an increased probability of traffic accidents \citep{chang2024safe, xie2024advdiffuser, xu2025diffscene}. These advanced foundation models offer strong controllability and exceptional environmental realism.

Despite their respective advantages in efficiency, exploration, or realism, these approaches share a fundamental limitation: their primary generation objectives are aligned with maximizing collision probabilities. Consequently, they can lead to practically unsolvable or action-indiscriminative situations, such as sudden rear-end collisions from blind spots \citep{mei2025seeking}. Training the driving agent on such unsolvable scenarios undermines the learning process, which severely degrades the overall task completion rate and driving efficiency in normal traffic conditions \citep{feng2026breaking}. 
Recent frameworks have attempted to mitigate unrealistic attacks by incorporating preference alignment \citep{nie2025steerable} or restricting adversarial behaviors within an estimated feasible region \citep{chen2024frea}. However, they do not explicitly align the generator with an ego-side resolvability signal within closed-loop training pipelines.
Therefore, there is a critical need to shift the focus of adversarial training from merely maximizing danger to ensuring the resolvability of the generated scenarios, ensuring that the driving agent is challenged by high-value critical scenarios.

\subsection{Scenario-based Curriculum Learning}
Effectively integrating generated scenarios into the training process is as crucial as the generation itself, where curriculum learning plays a vital role by organizing the training sequence to accelerate convergence and improve the learning process. While the principles of curriculum learning can be applied to various dimensions of the training pipeline, such as task decomposition \citep{qiao2018automatically, song2021autonomous} and reward shaping \citep{chu2025decision}, we focus on the schedule of scenarios. 
Early implementations of scenario-based curriculum learning primarily relied on manually designed heuristics or fixed stages, where scenarios are ordered based on static attributes such as traffic density \citep{anzalone2022end} or the difficulty score of the scenarios \citep{bronstein2023embedding}. 
To overcome the scalability and adaptability limitations of manual scheduling, recent research has shifted toward automated curriculum learning frameworks that dynamically adjust the training data distribution. For instance, \cite{sheng2026curricuvlm} progressively increases the probability of adversarial scenarios based on training phases. \cite{peng2024reward} models automated curriculum selection as a multi-armed bandit problem and utilizes historical rewards from reinforcement learning, thereby optimizing the sampling weights of each curriculum.
\cite{abouelazm2025automatic} dynamically selects scenarios by evaluating the learning potential of the driving policy, which is quantified by temporal-difference errors.

Despite the progress made by these automated frameworks, their scenario selection mechanisms remain fundamentally reactive. They predominantly rely on implicit difficulty estimates or reactive metrics derived from historical interactions. Since AD policy undergoes continuous and non-linear parameter updates, its ability to handle specific situations changes rapidly. Consequently, evaluating an agent based on its past performance often fails to reflect its true, current competence. They lack a forward-looking mechanism that explicitly predicts the agent's performance for given scenarios before interaction, leading to a potential mismatch between the selected scenarios and the policy's current capability \citep{xu2025heterogeneous}. Consequently, a predictive curriculum approach is essential for accurately matching training tasks with the policy's evolving capabilities.


\section{Preliminary and Problem Formulation}\label{sec:Preliminary}
\subsection{End-to-End Driving via RL}
The safe autonomous driving task can be formulated as a Markov Decision Process. We define this process using a tuple comprising the state space, the action space, the state transition dynamics, the reward function, and the discount factor, denoted as $\langle \mathcal{S}, \mathcal{A}, \mathcal{K}, \mathcal{R}, \gamma \rangle$. The environment is initialized from the scenario library denoted by $\Phi$. For a specific traffic scenario $\phi \in \Phi$, the state space $\mathcal{S}$ encompasses the kinematic states of the ego vehicle, the states of surrounding background vehicles, and the topological information of the road network. At each time step $t$, the autonomous driving agent observes the environment state $s_t \in \mathcal{S}$. Based on this observation, the agent executes a continuous control action $a_t \in \mathcal{A}$, which consists of longitudinal acceleration and lateral steering commands.

The state transition dynamics $\mathcal{K}: \mathcal{S} \times \mathcal{A} \rightarrow \Delta(\mathcal{S})$ describe the evolution of the traffic environment. The transition probability $\mathcal{K}(s_{t+1} | s_t, a_t, \phi)$ is jointly determined by the physical kinematics of the ego vehicle and the behavioral models of the background traffic participants within the specific scenario $\phi$. Then, the reward function $\mathcal{R}: \mathcal{S} \times \mathcal{A} \rightarrow \mathbb{R}$ provides a scalar feedback signal $r_t$ at each step, which is designed to encourage route completion while heavily penalizing safety violations such as collisions or road departures. Let $\pi_\theta: \mathcal{S} \rightarrow \mathcal{A}$ denote the end-to-end driving policy of the ego vehicle, parameterized by $\theta$. The fundamental objective of the autonomous driving agent is to learn an optimal policy $\pi_\theta^*$ that maximizes the expected cumulative discounted return over a finite time horizon $T$, which is formulated as:
\begin{equation}
    J(\pi_\theta, \phi) = \mathbb{E}_{a_t \sim \pi_\theta, s_{t+1} \sim \mathcal{K}} \left[ \sum_{t=0}^{T} \gamma^t \mathcal{R}(s_t, a_t) \right]
\end{equation}
where $J(\pi_\theta, \phi)$ denotes the expected cumulative discounted return of the driving policy $\pi_\theta$ under the specific scenario $\phi$, parameterized by the discount factor $\gamma$.

\subsection{Adversarial Training Objective}
Relying solely on naturalistic driving scenarios $\phi \in \Phi$ often fails to expose the autonomous driving agent to safety-critical events due to the long-tail distribution of traffic accidents. To address this rarity, the adversarial generator $\mathcal{G}_\psi$, parameterized by $\psi$, is introduced to perturb the original scenario $\phi$ into a safety-critical variant $\phi'$. By generating aggressive or challenging future trajectories for the critical background vehicles, the adversarial generator $\mathcal{G}_\psi$ acts as an attacker aiming to minimize the expected return of the current driving policy $\pi_\theta$. This adversarial scenario generation objective is formulated as:
\begin{equation}
\min_{\psi} \mathbb{E}_{\phi \sim \mathcal{P}(\phi), \phi' \sim \mathcal{G}_\psi(\cdot | \phi)} \left[ J(\pi_\theta, \phi') \right]
\end{equation}

Building upon adversarial scenario generation, closed-loop adversarial training establishes a two-player game to improve driving safety. The driving policy $\pi_\theta$ acts as the defender, aiming to learn robust strategies that maximize the expected return against the adversarial environments. The overall adversarial training paradigm thus yields a min-max optimization objective:
\begin{equation}
    \max_{\theta} \min_{\psi} \mathbb{E}_{\phi \sim \mathcal{P}(\phi), \phi' \sim \mathcal{G}_\psi(\cdot | \phi)} \left[ J(\pi_\theta, \phi') \right]
\end{equation}


However, most existing frameworks remain heavily attack-oriented. Directly optimizing this unconstrained min-max objective introduces two critical challenges that hinder the convergence and performance of the driving policy. 
First, an unconstrained adversarial generator $\mathcal{G}_\psi$ tends to generate practically unsolvable scenarios where unavoidable collisions occur. Under such conditions, the expected return $J(\pi_\theta, \phi')$ degenerates to a large negative penalty regardless of the agent's actions. This eliminates meaningful gradient signals for the policy, forcing it to collapse into overly conservative driving behaviors rather than learning effective accident-avoidance maneuvers.
Second, the training process usually relies on a scenario sampling mechanism, where a base scenario $\phi$ is drawn from the library $\Phi$ according to a sampling distribution $\mathcal{P}(\phi)$ at the onset of each training episode. Standard adversarial training assumes a static, uniform sampling distribution, formulated as $\mathcal{P}(\phi) = 1/|\Phi|$. As the policy $\pi_\theta$ continuously evolves during the training process, its capability to handle different scenarios fluctuates dynamically. Consequently, statically sampled scenarios may become trivially easy, leading to sample inefficiency and delayed convergence.

\subsection{Problem Formulation}
To address these bottlenecks, we reformulate the conventional adversarial training objective from a learning-oriented perspective. The goal is to convert attack-oriented adversarial scenarios into training scenarios that are both resolvable and informative for the evolving driving policy. Accordingly, we decouple the direct min-max optimization into two tractable sub-problems: resolvable adversarial scenario generation and predictive capability-aligned curriculum sampling. 

We constrain the adversarial generator $\mathcal{G}_\psi$ within a solvable space $\Omega_{\mathrm{solvable}}$. Formally, a scenario is defined as solvable if there exists at least one policy within the entire policy space $\Pi$ capable of safely navigating it. 
The objective of the scenario generator is thus redefined to maximize the risk against the current driving policy $\pi_\theta$ while satisfying the solvability constraint:
\begin{equation}
\label{equation_g}
\psi^{*}
=
\arg\min_{\psi}
\mathbb{E}_{\phi\sim \mathcal{P}(\phi),\ \phi'\sim G_{\psi}(\cdot\mid \phi)}
\left[
J(\pi_{\theta},\phi')
\right]
\quad
\mathrm{s.t.}\quad
\max_{\pi \in \Pi} J(\pi, \phi') \ge \delta_{\mathrm{safe}} 
\end{equation}
where $\delta_{\mathrm{safe}}$ denotes a predefined minimum safety threshold. By enforcing $\max_{\pi \in \Pi} J(\pi, \phi') \ge \delta_{\mathrm{safe}}$, we guarantee the existence of at least one solution, ensuring the generated scenarios are challenging yet solvable.

To enhance the training utility of the scenarios, we replace the static uniform sampling $\mathcal{P}(\phi)$ with a dynamic curriculum distribution $\mathcal{P}_{\theta, \omega}(\tilde{\phi})$. Here, $\tilde{\phi}$ denotes a generalized training scenario sampled from a mixed scenario library, which is either a naturalistic base scenario $\phi \in \Phi$ or a generated solvable adversarial scenario $\phi'$. This distribution is governed by a forward-looking policy capability prediction model $\mathcal{C}_\omega$, parameterized by $\omega$. The capability predictor explicitly estimates the probability $\mathcal{C}_\omega(\pi_\theta, \tilde{\phi}) \in [0, 1]$ that the current policy $\pi_\theta$ can successfully resolve the scenario $\tilde{\phi}$ prior to actual simulation. The dynamic curriculum distribution $\mathcal{P}_{\theta, \omega}(\tilde{\phi})$ is constructed to assign higher sampling weights to scenarios with lower predicted success probabilities, thereby prioritizing the exact vulnerabilities of the current policy and avoiding capability mismatch. Consequently, the final optimization objective for the driving policy is reformulated as:
\begin{equation}
    \theta^* = \arg\max_{\theta} \mathbb{E}_{\tilde{\phi} \sim \mathcal{P}_{\theta, \omega}(\tilde{\phi})} \left[ J(\pi_\theta, \tilde{\phi}) \right]
\end{equation}

\section{Methodology}\label{sec:Methodology}

\subsection{Framework Overview}
We propose a novel closed-loop adversarial training framework, denoted as AlignADV. The primary objective of this framework is to shift from the conventional unidirectional attack paradigm to learnability-guided curriculum loop, thereby accelerating the learning process and enhancing the ultimate robustness of the autonomous driving agent. Here, learnability refers to training utility of a scenario, which requires the scenario to be both practically resolvable and sufficiently challenging for current policy $\pi_\theta$. To achieve this objective, the framework couples resolvability-aligned scenario generation with capability-aligned sampling, so that the training scenarios are both practically resolvable and aligned with the evolving competence of the driving policy.

The AlignADV framework is constructed upon three synergistic modules. First, the \textbf{Solvable Adversarial Scenario Generation} module (Sec. \ref{sec:methodology-Generation}) optimizes the adversarial generator $\mathcal{G}_\psi$ to ensure that the generated safety-critical scenarios $\phi'$ reside within the solvable space $\Omega_{\mathrm{solvable}}$. Second, the \textbf{Dynamic Policy Capability Prediction} module (Sec. \ref{sec:methodology-Prediction}) constructs a parameterized predictor $\mathcal{C}_\omega$ to quantify the success probability of the current policy $\pi_\theta$ on any given scenario. In this module, we introduce a behavioral fingerprint to explicitly encode the intrinsic characteristics and dynamic responses of $\pi_\theta$. Third, the \textbf{Capability-Aligned Curriculum Sampling} module (Sec. \ref{sec:methodology-Sampling}) formulates the dynamic distribution $\mathcal{P}_{\theta, \omega}(\tilde{\phi})$ based on the outputs of $\mathcal{C}_\omega$, prioritizing the scenarios that maximize the learning gradient for the current policy.

The operational pipeline of the AlignADV framework is divided into two phases: an offline preparation phase and an online closed-loop training phase. During the offline phase, the adversarial generator $\mathcal{G}_\psi$ is fine-tuned via DPO, and the capability predictor $\mathcal{C}_\omega$ is trained based on the offline dataset. In the online policy training phase, policy optimization, scenario generation, and curriculum sampling are tightly integrated into a unified closed loop. As the driving policy $\pi_\theta$ evolves, the capability predictor $\mathcal{C}_\omega$ periodically evaluates the latest competence of the updated policy to construct the dynamic curriculum distribution $\mathcal{P}_{\theta, \omega}(\tilde{\phi})$. Meanwhile, the adversarial generator $\mathcal{G}_\psi$ continuously synthesizes resolvable safety-critical scenarios $\phi'$ tailored to the recent performance of the policy. Through this dynamic interaction, the framework continuously perceives the evolving competence of the driving policy to adaptively supply solvable, high-value challenges. 
This tightly coupled feedback loop ensures that the agent is neither overwhelmed by unsolvable hazards nor slowed down by trivial tasks, thus achieving highly efficient and robust autonomous driving training.


\begin{figure}
    \centering
    \includegraphics[width=0.95\linewidth]{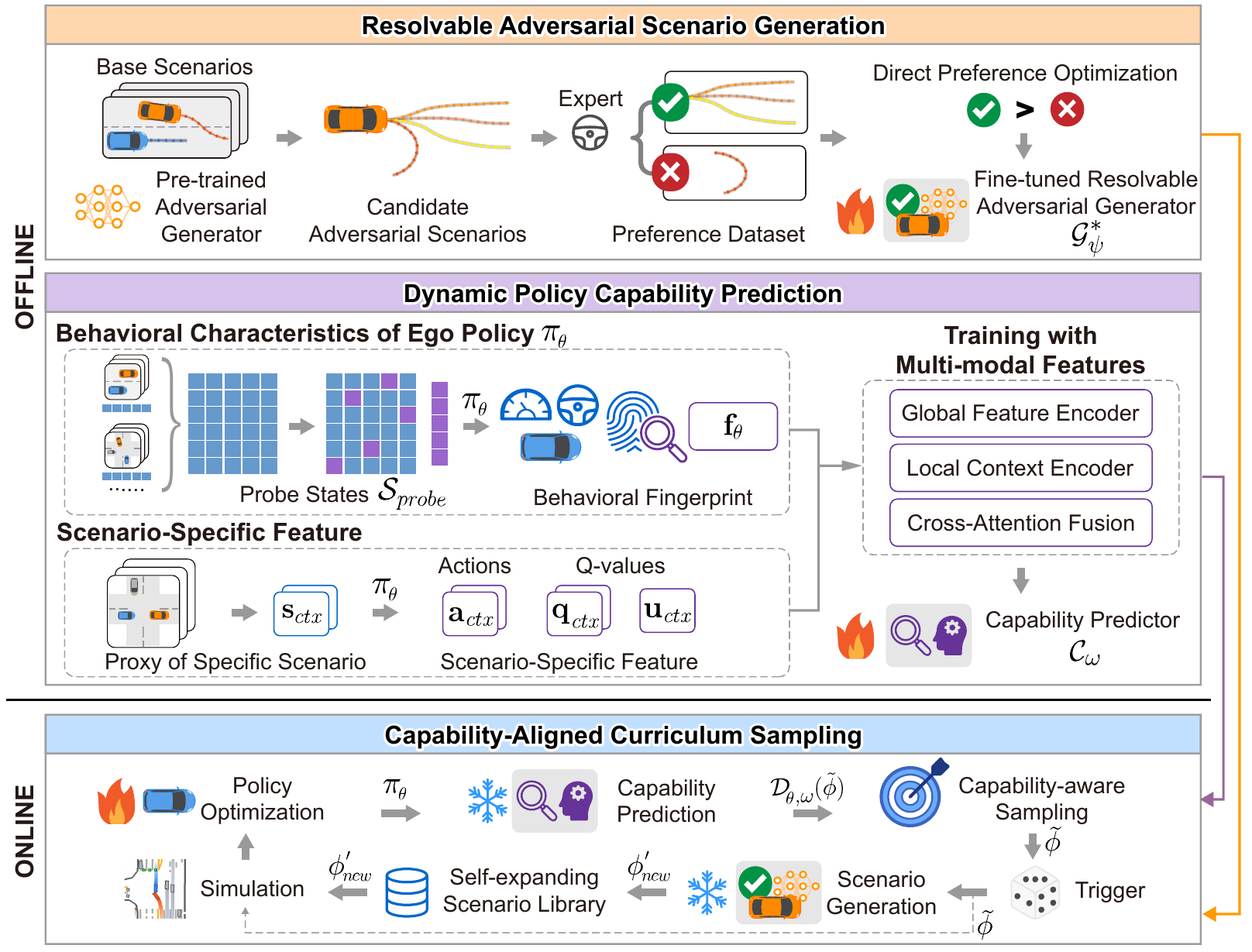}
    \caption{Overview of the AlignADV framework. (a) \textbf{Resolvable adversarial scenario generation} (Sec. \ref{sec:methodology-Generation}) fine-tunes a pretrained adversarial generator with expert-evaluated preference pairs and DPO, guiding generated scenarios toward critical yet resolvable interactions. (b) \textbf{Dynamic policy capability prediction} (Sec. \ref{sec:methodology-Prediction}) represents the evolving policy with behavioral fingerprints and predicts scenario-specific success probabilities before expensive simulation. (c) \textbf{Capability-Aligned curriculum sampling} (Sec. \ref{sec:methodology-Sampling}) combines the resolvable scenario library with predicted policy capabilities to construct a dynamic curriculum distribution for online policy optimization.}
    \label{fig:methodology_framework}
\end{figure}

\subsection{Resolvable Scenario Generation via Preference Optimization} \label{sec:methodology-Generation} 
The mechanism of the adversarial generator $\mathcal{G}_\psi$ is to synthesize a safety-critical scenario $\phi'$ by manipulating the future trajectory of the background vehicle, which interacts with the ego vehicle. A naturalistic base scenario $\phi \in \Phi$ encompasses the static map $\mathcal{M}$ alongside the states of all traffic participants. To predict the background vehicle's future behavior, the generator extracts a historical scenario tuple $\phi_{his} = (\mathcal{M}, \mathbf{s}_{ego}, \mathbf{s}_{bv})$ from $\phi$, where $\mathbf{s}_{ego}$ and $\mathbf{s}_{bv}$ denote the historical state sequences of the ego vehicle and the background vehicles. Built upon a deep learning motion forecasting model, the generator takes this scenario tuple as input to predict a multi-modal trajectory distribution for the critical background vehicle, turning it into the adversarial vehicle. From the distribution, a set of $K$ candidate trajectories $\mathcal{T}_\phi$ is sampled, which can be denoted as:
\begin{equation}
    \mathcal{T}_\phi = \left\{ \tau_{adv}^{(k)} \sim \mathcal{G}_\psi(\cdot \mid \phi_{his}) \right\}_{k=1}^K
\end{equation}
Subsequently, each candidate trajectory $\tau_{adv}^{(k)} \in \mathcal{T}_\phi$ is integrated into the base scenario $\phi$, replacing its original benign trajectory, to construct a perturbed adversarial scenario $\phi'_k = \phi \cup \{ \tau_{adv}^{(k)} \}$. 

To construct an effective training environment, the optimal adversarial trajectory $\tau_{adv}^*$ must be selected to maximize the collision likelihood against the current driving policy $\pi_\theta$ while residing within the solvable space $\Omega_{\mathrm{solvable}}$. Following the probabilistic factorization principle \citep{zhang2023cat}, the joint collision probability can be decomposed into the traffic prior, the ego vehicle estimation, and the collision likelihood, which is mathematically formulated as:
\begin{equation}
    \tau_{adv}^* = \arg\max_{\tau_{adv} \in \mathcal{T}_\phi} \mathbb{P}(\tau_{adv}|\phi_{his}) \mathbb{P}(\tau_{ego}|\tau_{adv}, \phi_{his}, \pi_\theta) \mathbb{P}(Coll|\tau_{ego}, \tau_{adv}) \quad \text{s.t.} \quad \phi\cup\{\tau_{\mathrm{adv}}\}\in \Omega_{\mathrm{solvable}}
\end{equation}

where $\tau_{ego}$ denotes the future trajectory of the ego vehicle rolled out by the policy $\pi_\theta$, and $Coll$ denotes the collision event. The constraint $\phi\cup\{\tau_{\mathrm{adv}}\}\in \Omega_{\mathrm{solvable}}$ ensures that every generated candidate trajectory provides a feasible solution for the ego vehicle. Concurrently, the collision maximization objective encourages these solvable scenarios to effectively expose the weaknesses of the current policy $\pi_\theta$, forcing the agent to learn robust accident-avoidance strategies.


To systematically quantify the solvability of the generated candidate trajectories, we introduce a privileged rule-based expert policy, denoted as $\pi_{expert}$, equipped with perfect environmental perception and precise kinematic control. Based on the ground truth future states of the background vehicles, the privileged expert policy employs a kinematic bicycle model combined with proactive proportional-integral-derivative (PID) controllers to navigate the ego vehicle. More details about the expert policy are provided in \ref{sec:Appendix-expertpolicy}. For each perturbed scenario $\phi'_k$, we conduct a closed-loop rollout by resetting the simulator with the generated adversarial trajectory and using $\pi_{expert}$ as the ego-vehicle controller.
Then, we define an evaluation function $\mathcal{E}(\pi_{expert}, \phi'_k)$ that yields two critical metrics including the binary collision indicator $\mathcal{I}_{col} \in \{0, 1\}$ and the continuous route completion rate $\mathcal{I}_{rc} \in [0, 1]$. Based on these rollout metrics, we define an expert-based solvability indicator $\mathcal{I}_{solvable}(\tau_{adv}^{(k)})$ for a given trajectory as:
\begin{equation}
    \mathcal{I}_{solvable}(\tau_{adv}^{(k)}) = 
    \begin{cases} 
        1, & \text{if } \mathcal{I}_{col} = 0 \text{ and } \mathcal{I}_{rc}  \ge \eta_{rc}, \text{ where } (\mathcal{I}_{col}, \mathcal{I}_{rc}) = \mathcal{E}(\pi_{expert}, \phi'_k) \\
        0, & \text{otherwise}
    \end{cases}
\end{equation}
where $\eta_{rc}$ represents a predefined threshold for acceptable route completion. This indicator explicitly determines whether a generated trajectory resides within the theoretical solvable space $\Omega_{\mathrm{solvable}}$. 

Using this indicator, the candidate trajectory set $\mathcal{T}_\phi$ is partitioned into a solvable subset $\mathcal{T}_{sol}$ and an unsolvable subset $\mathcal{T}_{unsol}$:
\begin{equation}
\begin{aligned}
    \mathcal{T}_{sol} &= \left\{ \tau_{adv}^{(k)} \in \mathcal{T}_\phi \mid \mathcal{I}_{solvable}(\tau_{adv}^{(k)}) = 1 \right\} \\
    \mathcal{T}_{unsol} &= \left\{ \tau_{adv}^{(k)} \in \mathcal{T}_\phi \mid \mathcal{I}_{solvable}(\tau_{adv}^{(k)}) = 0 \right\}
\end{aligned}
\end{equation}

Based on this deterministic evaluation, we construct a preference dataset $\mathcal{D}_{pref}$ to guide the generator $\mathcal{G}_\psi$ towards the solvable space. For a given base scenario $\phi$, we formulate preference pairs $\tau_w$ and $\tau_l$, where $\tau_w$ denotes the preferred or winner trajectory, $\tau_l$ represents the dispreferred or loser trajectory, and $N_{\mathrm{pref}}=|\mathcal{D}_{\mathrm{pref}}|$ denotes the total number
of preference pairs collected from all eligible base scenarios.
\begin{equation}
\mathcal{D}_{\mathrm{pref}}
=
\left\{
\left(
\phi_{\mathrm{his}}^{(n)},
\tau_w^{(n)},
\tau_l^{(n)}
\right)
\right\}_{n=1}^{N_{\mathrm{pref}}}
\end{equation}

When both the solvable and unsolvable subsets are non-empty, we construct the preference pairs by uniformly sampling $\tau_w \sim \mathcal{T}_{sol}$ and $\tau_l \sim \mathcal{T}_{unsol}$. Scenarios where all candidate trajectories are solvable are excluded from the preference dataset to concentrate the optimization efforts on the boundary of safety-critical failures. Conversely, under highly aggressive scenario initializations where all candidate trajectories are evaluated as unsolvable, we construct pseudo-preference pairs based on the continuous route completion metric $\mathcal{I}_{rc}$ to provide dense learning signals. Specifically, a trajectory $\tau_i$ is preferred over $\tau_j$ if $\mathcal{I}_{rc}(\tau_i) > \mathcal{I}_{rc}(\tau_j) + \epsilon$, where $\epsilon$ is a marginal threshold. The threshold $\epsilon$ ensures a meaningful difference in the driving progress, indicating that $\tau_i$ allows the ego vehicle to survive longer before the inevitable failure occurs.

With the preference dataset $\mathcal{D}_{pref}$ established, we employ the DPO \citep{rafailov2023direct} algorithm to fine-tune the parameters $\psi$ of the adversarial generator $\mathcal{G}_\psi$. Unlike traditional reinforcement learning from human feedback that requires training an explicit reward model, DPO directly optimizes the network to align with the established preferences. Let $\mathcal{G}_{\psi_{ref}}$ denote the frozen pre-trained reference model, which represents the original unconstrained trajectory forecasting model without preference alignment. The generator $\mathcal{G}_\psi$ is updated by minimizing the negative log-sigmoid of the implicit reward difference between the winner and loser trajectories. The optimization objective is formally defined as:
\begin{equation}
    \mathcal{L}_{DPO}(\psi) = -\mathbb{E}_{(\phi_{his}, \tau_w, \tau_l) \sim \mathcal{D}_{pref}} \left[ \log \sigma \left( \beta \log \frac{\mathcal{G}_\psi(\tau_w|\phi_{his})}{\mathcal{G}_{\psi_{ref}}(\tau_w|\phi_{his})} - \beta \log \frac{\mathcal{G}_\psi(\tau_l|\phi_{his})}{\mathcal{G}_{\psi_{ref}}(\tau_l|\phi_{his})} \right) \right]
\end{equation}
where $\sigma$ is the sigmoid function and $\beta$ is a temperature hyperparameter controlling the deviation from the reference model. 

By minimizing $\mathcal{L}_{DPO}(\psi)$, the generator intrinsically increases the likelihood of synthesizing solvable trajectories $\tau_w$ while penalizing the probability of generating unsolvable trajectories $\tau_l$. Through this offline fine-tuning process, the optimized adversarial generator, now denoted as $\mathcal{G}_{\psi}^*$, is constrained within the solvable space $\Omega_{\mathrm{solvable}}$. This satisfies the theoretical generation objective defined in Eq.\ref{equation_g}, thereby providing high-quality, solvable safety-critical environments for the subsequent closed-loop policy training.

\subsection{Dynamic Policy Capability Prediction} \label{sec:methodology-Prediction}
To implement a highly efficient curriculum learning paradigm, it is essential to dynamically evaluate the evolving capability of the current policy, which is concretely reflected by its success probability on any given scenario $\tilde{\phi}$. 
However, the driving policy is a highly non-linear neural network whose competence fluctuates dynamically during the training process. Furthermore, the traffic scenarios exhibit immense spatial-temporal diversity. Directly evaluating this success probability by executing full closed-loop simulations on every candidate scenario is computationally prohibitive. 
To tackle these challenges, we propose a parameterized capability predictor $\mathcal{C}_\omega$. The core mechanism of this predictor is to map the intrinsic behavioral characteristics of the current policy $\pi_\theta$ and its localized responses within a specific scenario $\tilde{\phi}$ to a continuous success probability. The prediction framework is structurally divided into two components, including the extraction of a scenario-agnostic behavioral characteristics and the construction of a multi-modal capability prediction network.

\subsubsection{Behavioral Fingerprint Extraction}
We introduce the concept of \textbf{behavioral fingerprint} to represent the intrinsic driving style and decision-making preferences of the evolving driving policy. As the driving policy undergoes continuous parameter updates, its strategic mapping from the state space to the action space, formally denoted as $\pi_\theta: \mathcal{S} \rightarrow \mathcal{A}$, shifts dynamically. The behavioral fingerprint captures this mapping by calculating the deterministic actions of the policy across a meticulously designed set of standardized probe states. Let $\mathcal{S}_{probe}$ denote the probe state set comprising $M$ states: 
\begin{equation}
    \mathcal{S}_{probe} = \{s_{probe}^{(1)}, s_{probe}^{(2)}, \dots, s_{probe}^{(M)}\} \subset \mathcal{S}
\end{equation}

The construction of $\mathcal{S}_{probe}$ is conducted offline by aggregating a massive state dataset $\mathcal{D}_{state}$ derived from naturalistic scenarios. To ensure that the probe states comprehensively cover both typical driving conditions and safety-critical boundary cases, we employ a hybrid extraction methodology. First, we apply the K-Means clustering algorithm to the state dataset $\mathcal{D}_{state}$, extracting the cluster centroids to form a representative state subset $\mathcal{S}_{rep}$. Second, to ensure the inclusion of safety-critical boundaries, we evaluate the Time-to-Collision (TTC) metric for all states in $\mathcal{D}_{state}$ and heuristically filter a critical state subset $\mathcal{S}_{crit}$, characterized by minimal TTC values. The final probe state set is formulated by the union and de-duplication of these two subsets, denoted as $\mathcal{S}_{probe} = \mathcal{S}_{rep} \cup \mathcal{S}_{crit}$.

For any given policy $\pi_\theta$ parameterized by $\theta$ at a specific training step, the behavioral fingerprint $\mathbf{f}_\theta$ is generated by computing the action vectors across the entire probe state set $\mathcal{S}_{probe}$. Specifically, for each probe state $s_{probe}^{(m)}$, the policy outputs the deterministic continuous control action $a^{(m)} = \pi_\theta(s_{probe}^{(m)}) \in \mathcal{A}$, containing both longitudinal acceleration and lateral steering commands. The behavioral fingerprint $\mathbf{f}_\theta$ of the given policy $\pi_\theta$ is then defined as the concatenation of all corresponding action vectors:
\begin{equation}
    \mathbf{f}_\theta = \bigoplus_{m=1}^{M} \pi_\theta \left( s_{probe}^{(m)} \right)
\end{equation}
where $\bigoplus$ denotes the concatenation operator. Given that $\mathcal{A} \subset \mathbb{R}^2$, this yields a vector $\mathbf{f}_\theta \in \mathbb{R}^{2M}$ that acts as a unique signature of the policy, providing the subsequent prediction model with a global representation of the algorithmic competence.

\subsubsection{Scenario-Specific Feature Construction}
While the behavioral fingerprint provides a global representation of the policy $\pi_\theta$, explicitly estimating the success probability $\mathcal{C}_\omega(\pi_\theta, \tilde{\phi})$ requires localized interaction contexts within the specific scenario $\tilde{\phi}$. 
To achieve this, we formulate the capability predictor $\mathcal{C}_\omega$ as a multi-modal deep neural network incorporating a cross-attention Transformer architecture. The model integrates the global behavioral fingerprint $f_\theta$ with a sequence of localized interaction features. 

To accurately predict the policy performance without executing a computationally expensive full closed-loop simulation, we must construct a highly representative proxy of the specific scenario $\tilde{\phi}$. 
The driving policy changes incrementally between consecutive training updates. Therefore, instead of executing a new rollout for every candidate scenario, we reuse the historical states encountered by the previous policy as a close proxy for the situations that the current policy $\pi_\theta$ is likely to face in the same scenario.
Specifically, we extract a fixed-length context state sequence $\mathbf{s}_{ctx} \in \mathbb{R}^{L \times |\mathcal{S}|}$ from the historical states. To maximize information density, the sequence is constructed based on a hybrid sampling strategy:  
\begin{equation}
    \mathbf{s}_{ctx} = (s_{t_1}, s_{t_2}, \dots, s_{t_L}), \quad \text{where } \{t_1, \dots, t_L\} = \mathcal{F}_{init} \cup \mathcal{F}_{crit} \cup \mathcal{F}_{sparse}   
\end{equation}
where $\mathcal{F}_{init}$ represents the initial frames to capture the scenario initialization, $\mathcal{F}_{crit}$ selects the critical frames with the minimum TTC to capture the core conflict, and $\mathcal{F}_{sparse}$ uniformly samples the remaining frames to represent the overall progression. Unlike the scenario-agnostic probe set $\mathcal{S}_{probe}$, the sequence $\mathbf{s}_{ctx}$ explicitly defines the localized environmental challenges the current policy needs to resolve within $\tilde{\phi}$.

Merely observing the states is insufficient to predict whether the current policy can safely complete the scenario. The predictor must evaluate how the current policy intends to react to these hazards and assess the quality of those reactions. 
First, we query the current policy $\pi_\theta$ using $\mathbf{s}_{ctx}$ to obtain the corresponding action sequence $\mathbf{a}_{ctx} \in \mathbb{R}^{L \times |\mathcal{A}|}$. 
Second, to rigorously evaluate the safety and confidence of these actions, we incorporate multidimensional value functions. 
For each time step $t_l$, we compute the subjective Q-values $(Q_{\theta,1}, Q_{\theta,2})$ generated by the twin critics of the current policy, which reflect its internal confidence and uncertainty. 
However, subjective Q-values are prone to overestimation during the early stages of reinforcement learning. Therefore, we introduce an objective baseline by querying two pre-trained expert policies $\pi_{E_1}, \pi_{E_2}$, each equipped with twin critics. The comprehensive Q-value feature vector for a specific state-action pair is concatenated as:
\begin{equation}
    q_{t_l} = \left[ Q_{\theta,1}(s_{t_l}, a_{t_l}), Q_{\theta,2}(s_{t_l}, a_{t_l}), Q_{E_1,1}(s_{t_l}, a_{t_l}),Q_{E_1,2}(s_{t_l}, a_{t_l}), Q_{E_2,1}(s_{t_l}, a_{t_l}), Q_{E_2,2}(s_{t_l}, a_{t_l}) \right]^T \in \mathbb{R}^6
\end{equation}

These vectors form the sequential Q-value feature $\mathbf{q}_{ctx} = (q_{t_1}, \dots, q_{t_L}) \in \mathbb{R}^{L \times 6}$. Moreover, a statistical feature vector $\mathbf{u}_{ctx} \in \mathbb{R}^{24}$ comprising the mean, maximum, minimum, and standard deviation of these Q-values is computed to provide robust scalar indicators of the overall scenario risk.

\subsubsection{Capability Prediction Model}
With the multi-modal features constructed, the capability predictor $\mathcal{C}_\omega$ processes these inputs through a hierarchical attention-based architecture. 

The localized interaction features, including the actions $\mathbf{a}_{ctx}$ and Q-values $\mathbf{q}_{ctx}$, are projected into a latent space via multi-layer perceptrons. These latent representations are then concatenated along the feature dimension to form a unified sequential embedding $\mathbf{X}_{seq} \in \mathbb{R}^{L \times d_{model}}$. 
\begin{equation}
    \mathbf{X}_{seq} = \text{MLP}_{a}(\mathbf{a}_{ctx}) \oplus \text{MLP}_{q}(\mathbf{q}_{ctx})
\end{equation}
where $\oplus$ denotes the concatenation operator, and each MLP consists of a linear transformation, a ReLU activation, and dropout regularization. 

A multi-head self-attention mechanism \citep{NIPS2017_3f5ee243}, augmented with positional encoding ($\text{PE}$) and a residual connection, is subsequently applied to model the temporal dependencies within the interaction sequence:

\vspace{-4pt}
\begin{equation}
    \mathbf{X}_{PE} = \mathbf{X}_{seq} + \text{PE}
\end{equation}
\begin{equation}
    \mathbf{E}_{seq} = \mathbf{X}_{PE} + \text{Dropout} \left( \text{softmax}\left( \frac{(\mathbf{X}_{PE}\mathbf{W}^Q)(\mathbf{X}_{PE}\mathbf{W}^K)^T}{\sqrt{d_{model}}} \right) (\mathbf{X}_{PE}\mathbf{W}^V) \right)
\end{equation}
where $\mathbf{W}^Q,\mathbf{W}^K,\mathbf{W}^V \in \mathbb{R}^{d_{model} \times d_{model}}$ are the learnable linear projection matrices for queries, keys, and values, respectively.

The global features, comprising the behavioral fingerprint $\mathbf{f}_\theta$ and the Q-statistics $\mathbf{u}_{ctx}$, are encoded into a global embedding $\mathbf{E}_{global} \in \mathbb{R}^{d_{model}}$. 
\begin{equation}
    \mathbf{E}_{global} = \mathbf{W}_{proj} \Big( \text{MLP}_{f}(\mathbf{f}_\theta) \oplus \text{MLP}_{u}(\mathbf{u}_{ctx}) \Big) + \mathbf{b}_{proj}
\end{equation}
where $\mathbf{W}_{proj}$ and $\mathbf{b}_{proj}$ are the learnable weight matrix and bias vector of the global projection layer.

To effectively fuse the global features with the localized scenario context, we employ a cross-attention mechanism. The global embedding $\mathbf{E}_{global}$ serves as the query, while the temporally encoded sequential embedding $\mathbf{E}_{seq}$ serves as both the key and the value. 
\begin{equation}
    \mathbf{H}_{fusion} = \text{softmax}\left( \frac{(\mathbf{E}_{global}\mathbf{W}_{cross}^Q)(\mathbf{E}_{seq}\mathbf{W}_{cross}^K)^T}{\sqrt{d_{model}}} \right) (\mathbf{E}_{seq}\mathbf{W}_{cross}^V)
\end{equation}
where $\mathbf{W}^Q_{cross},\mathbf{W}^K_{cross},\mathbf{W}^V_{cross} \in \mathbb{R}^{d_{model} \times d_{model}}$ are the learnable projection matrices for the cross-attention mechanism.

This architecture enables the global representation of the agent to dynamically attend to the most critical moments within the scenario sequence. 
Ultimately, the estimated success probability, denoted as $\hat{p} \in [0, 1]$, is realized by processing the cross-attention output through a fully connected prediction head with a sigmoid activation function:
\begin{equation}
     \hat{p} = \mathcal{C}_\omega(\pi_\theta, \tilde{\phi}) := \mathcal{C}_\omega(\mathbf{f}_\theta, \mathbf{a}_{ctx}, \mathbf{q}_{ctx}, \mathbf{u}_{ctx}) = \sigma \big( \text{MLP} (\mathbf{H}_{fusion}) \big)
\end{equation}

The parameters $\omega$ of the capability predictor are optimized through an offline supervised pre-training paradigm prior to the online adversarial training of the driving policy. We first run independent closed-loop simulations with driving policies saved at different learning stages. For each saved policy, we roll it out over scenarios and record the encountered states, policy actions, value estimates, and episode outcomes. These policy-scenario interaction records form an offline supervised dataset for capability prediction.
Specifically, the scenario context $\mathbf{s}_{ctx}$ is derived from the rollout of the earlier policy $\pi_{i-1}$, while the behavioral fingerprint $\mathbf{f}_\theta$, current actions $\mathbf{a}_{ctx}$, and the ground-truth success label $y \in \{0, 1\}$ are derived from the subsequent policy $\pi_i$. The predictor is optimized using the binary cross-entropy loss between the predicted probability $\hat{p}$ and the actual success label $y$:
\begin{equation}
    \mathcal{L}_{pred}(\omega) = -\mathbb{E} \left[ y \log(\hat{p}) + (1 - y) \log(1 - \hat{p}) \right]
\end{equation}
Once trained, the predictor $\mathcal{C}_\omega$ is frozen and deployed in the online adversarial training loop to dynamically evaluate the evolving policy and guide the curriculum sampling distribution.

\subsection{Capability-Aligned Curriculum Sampling} \label{sec:methodology-Sampling}
Building upon the resolvability-aligned adversarial generator $\mathcal{G}_\psi$ and the dynamic capability predictor $\mathcal{C}_\omega$, we formulate a capability-aligned curriculum learning paradigm driven by the mutual evolution of the autonomous driving agent and the training environment.
We first establish a \textbf{self-expanding scenario library}, formulated as $\tilde{\Phi}_t = \Phi \cup \Phi_{adv}^{(t)}$. Here, $\Phi$ represents the static base scenario library and $\Phi_{adv}^{(t)}$ denotes the set of generated adversarial scenarios dynamically maintained at training step $t$. 
As the driving policy $\pi_\theta$ continuously updates, newly generated adversarial scenarios are iteratively incorporated into the library to progressively scale the environmental difficulty. 
Then, to efficiently navigate this evolving environment, we dynamically construct a capability-aligned sampling distribution $\mathcal{P}_{\theta, \omega}(\tilde{\phi})$. Instead of uniformly sampling, our proposed framework leverages the predicted success probability $\hat{p}$ to prioritize scenarios that expose the current vulnerabilities of the evolving policy $\pi_\theta$ from the self-expanding library $\tilde{\Phi}_t$.


To construct the capability-aligned sampling distribution $\mathcal{P}_{\theta, \omega}(\tilde{\phi})$, we introduce a difficulty-proportional weighting mechanism.
For any candidate scenario $\tilde{\phi} \in \tilde{\Phi}_t$, the capability predictor outputs a continuous success probability $\hat{p}_{\tilde{\phi}}$ of the current driving policy $\pi_\theta$.
To prioritize the exact vulnerabilities of the policy, the capability-aligned weighting function $\mathcal{W}(\tilde{\phi})$ is formulated as the complement of the predicted success probability.
By assigning higher sampling probabilities to scenarios where the agent is most likely to fail, the framework explicitly forces the policy to concentrate on hard scenarios rather than wasting computational resources on trivially solvable tasks. 
The sampling weight for a specific scenario is defined as:
\begin{equation}
    \mathcal{W}(\tilde{\phi}) = 1 - \mathcal{C}_\omega(\pi_\theta, \tilde{\phi})
\end{equation}

Consequently, the dynamic curriculum sampling distribution at any given training stage is formalized by normalizing these weights across the entire self-expanding library. Through the synergy of capability-aligned sampling weighting and dynamic library expansion, the training environment continuously strengthens in response to the evolving capability of the driving policy.
\begin{equation}
    \mathcal{P}_{\theta, \omega}(\tilde{\phi}) = \frac{\mathcal{W}(\tilde{\phi})}{\sum_{\tilde{\phi}_k \in \tilde{\Phi}_t} \mathcal{W}(\tilde{\phi}_k)}
\end{equation}

In practical implementations, executing a global prediction across the entire scenario library at every single training step requires substantial computational costs and leads to unnecessary resource waste. To optimize computational efficiency, the capability prediction and the subsequent distribution update are executed periodically at a fixed interval of $N_{eval}$ timesteps. Between two consecutive prediction intervals, the capability-aligned sampling distribution $\mathcal{P}_{\theta, \omega}(\tilde{\phi})$ is frozen.

At the onset of each training episode, the framework determines whether to generate a new adversarial scenario using a time-varying trigger threshold. To gradually increase the frequency of adversarial scenario generation as the policy matures, we define the trigger threshold as:
\begin{equation}
    \rho_t = \max \left( \rho_{min}, 1 - \kappa \frac{t}{T_{max}} (1 - \rho_{min}) \right),
\end{equation}
where $\rho_{min}$ denotes the lower bound of the trigger threshold, $\kappa$ controls the decay rate, and $T_{max}$ represents the maximum training horizon.

A uniform random variable $u_t \sim \mathcal{U}(0,1)$ is then sampled, and the binary generation indicator is defined as:
\begin{equation}
    g_t = \mathbb{I}(u_t > \rho_t).
\end{equation}
Accordingly, the actual probability of generating an adversarial scenario is:
\begin{equation}
    \Pr(g_t=1) = 1 - \rho_t.
\end{equation}



When $g_t = 1$, the generation mechanism is triggered, and the fine-tuned adversarial generator $\mathcal{G}_{\psi}^*$ is activated to perturb the selected scenario, yielding a newly solvable adversarial scenario $\phi'$. The generated scenario is subsequently appended to the dynamic library $\tilde{\Phi}_t$. Conversely, when $g_t = 0$, the agent directly trains on the currently sampled scenario without perturbation. To prevent unbounded library expansion and maintain the freshness of the adversarial challenges, we impose a strict capacity limit on the self-expanding library. Specifically, each base scenario $\phi \in \Phi$ is permitted to maintain a maximum of $N_{max}$ associated adversarial variants. When the generation indicator triggers the synthesis of a new variant and the corresponding base scenario has already reached this capacity limit, a prioritized chronological eviction policy is executed to permanently remove the oldest adversarial variant. This continuous influx of newly generated adversarial scenarios, coupled with the capability-aligned sampling prioritization and strict capacity management, forms a robust closed-loop curriculum. 

The complete training paradigm of the AlignADV framework, integrating the offline preference optimization, the capability predictor pre-training, and the online capability-aligned curriculum learning, is systematically summarized in Algorithm \ref{alg:framework}.

\begin{algorithm}[!htbp]
\caption{AlignADV: Learnability-guided Adversarial Training Framework}
\label{alg:framework}
\small
\KwIn{Base scenario library $\Phi$, pre-trained reference generator $\mathcal{G}_{\psi_{ref}}$, Offline state dataset $\mathcal{D}_{state}$, Pre-trained expert critic $Q_{E_1}, Q_{E_2}$, Maximum training steps $T_{max}$, Evaluation interval $N_{eval}$, Capacity limit $N_{max}$.}
\KwOut{Optimal autonomous driving policy $\pi_\theta^*$}
\tcp{Phase 1: Offline Preparation}
Construct preference dataset $\mathcal{D}_{pref}$ using the rule-based expert policy $\pi_E$\;
Fine-tune generator $\mathcal{G}_\psi \leftarrow \arg\min_\psi \mathcal{L}_{DPO}(\psi)$ using $\mathcal{D}_{pref}$ via DPO\;
Construct probe state set \(S_{probe}\) via K-Means and TTC analysis on $\mathcal{D}_{state}$\;
Train capability predictor $\mathcal{C}_\omega \leftarrow \arg\min_\omega \mathcal{L}_{pred}(\omega)$ using multi-modal features\;
\vspace{0.05cm}
\tcp{Phase 2: Online Closed-Loop Training}
Initialize RL policy $\pi_\theta$ and replay buffer $\mathcal{B}$\;
Initialize dynamic scenario library $\tilde{\Phi}_0 \leftarrow \Phi$\;
Initialize evaluation marker \(t_{\mathrm{eval}}\leftarrow 0\)\;
\While{total training step $t < T_{max}$}{
    \tcp{Predictor Evaluation and Sampling Weight Updating}
    \If{$t - t_{eval} \ge N_{eval}$}{
        Extract behavioral fingerprint $\mathbf{f}_\theta$ for current policy $\pi_\theta$ using $\mathcal{S}_{probe}$\;
        \For{each scenario $\tilde{\phi}_i \in \tilde{\Phi}_t$}{
            Extract historical context state sequence $\mathbf{s}_{ctx}$ from $\tilde{\phi}_i$\;
            Query current policy to obtain intended actions: $\mathbf{a}_{ctx} \leftarrow \pi_\theta(\mathbf{s}_{ctx})$\;
            Use critics to obtain Q-values: $\mathbf{q}_{ctx} \leftarrow Q_\theta(\mathbf{s}_{ctx}, \mathbf{a}_{ctx}) \cup Q_E(\mathbf{s}_{ctx}, \mathbf{a}_{ctx})$\;
            Compute statistical features $\mathbf{u}_{ctx}$ from $\mathbf{q}_{ctx}$\;
            Predict success probability $\hat{p}_i \leftarrow \mathcal{C}_\omega(\mathbf{f}_\theta, \mathbf{a}_{ctx}, \mathbf{q}_{ctx}, \mathbf{u}_{ctx})$\;
        }
        Compute weights $\mathcal{W}(\tilde{\phi}_i) = 1 - \hat{p}_i$ and update distribution $\mathcal{P}_{\theta, \omega}(\tilde{\phi})$\;
        Update evaluation marker: $t_{eval} \leftarrow t$\;
    }
    Sample a training scenario $\tilde{\phi} \sim \mathcal{P}_{\theta, \omega}(\tilde{\phi})$\;
    \tcp{Adversarial Scenario Generation}
    Sample binary generation indicator $g_t = \mathbb{I}(u_t > \rho_t).$\;
    \If{$g_t == 1$}{
        Synthesize solvable adversarial scenario $\phi'_{new} \sim \mathcal{G}_{\psi}^*(\cdot | \tilde{\phi})$\;
        \If{capacity threshold $N_{max}$ is exceeded for the base scenario}{
            Remove the oldest adversarial variant $\phi'_{old}$ from $\tilde{\Phi}_t$\;
        }
        Append the newly generated scenario: $\tilde{\Phi}_t \leftarrow \tilde{\Phi}_t \cup \{\phi'_{new}\}$\;
        Replace current training scenario: $\tilde{\phi} \leftarrow \phi'_{new}$\;
    }
    \tcp{RL Optimization}
    Reset environment with scenario $\tilde{\phi}$ and execute episode using policy $\pi_\theta$\;
    Collect transition tuples $(s, a, r, s')$ into replay buffer $\mathcal{B}$ and record episode steps $\Delta t$\;
    Update policy parameters $\theta$ via standard RL optimization\;
    Accumulate total steps: $t \leftarrow t + \Delta t$\;
    }
\Return{$\pi_\theta^*$}
\end{algorithm}

\section{Experiment}\label{sec:experiments}
This section designs comprehensive experiments to evaluate the proposed learnability-guided closed-loop training framework. We first verify the quality of the generated scenarios by analyzing solvability and diversity metrics. We then visualize the extracted behavior fingerprints and evaluate the accuracy of the capability prediction model through comprehensive ablation studies. Finally, we present the adversarial reinforcement learning results by comparing the proposed framework against adversarial training and sampling strategy baselines.
\subsection{Experimental Setup}
All experiments are conducted utilizing the MetaDrive simulator \citep{li2022metadrive}, a lightweight and efficient simulation platform for autonomous driving. We use this simulator to reconstruct real-world traffic logs sourced from the Waymo Open Motion Dataset (WOMD) \citep{ettinger2021large}. This reconstruction allows for closed-loop evaluation, enabling the ego agent to react dynamically to the behaviors of traffic participants. A subset of 480 scenarios from the WOMD dataset is utilized to construct the experimental environment. These scenarios are divided into a training set of 382 scenarios and an independent testing set of 98 scenarios. To ensure high-fidelity vehicle kinematics and realistic traffic interactions, the simulation operates at a control frequency of 10 Hertz. Furthermore, each scenario spans a total temporal horizon of 9 seconds, which incorporates 1 second of historical trajectory data.

The autonomous driving policy is trained using the Twin Delayed Deep Deterministic Policy Gradient algorithm (TD3). For the scenario generation module, we employ DenseTNT as the pretrained reference generator. The parameters of TD3 and DenseTNT are defined following the setup in \cite{zhang2023cat} and \cite{nie2025steerable}.

The implementation of the proposed AlignADV framework necessitates the configuration of several critical hyperparameters. The precise numerical values defined in the solvable scenario generation, dynamic capability prediction, and capability-aligned curriculum sampling modules are systematically summarized in Table \ref{tab:hyperparameters}.

\begin{table}[htbp]
    \centering
    \small
    \caption{Summary of critical hyperparameters for the AlignADV framework.}
    \label{tab:hyperparameters}
    \renewcommand{\arraystretch}{1.15}
    \setlength{\tabcolsep}{12pt}
    \begin{tabular}{llc}
        \toprule
        \textbf{Module} & \textbf{Hyperparameter} & \textbf{Value} \\
        \midrule
        \multirow{4}{*}{Solvable Generation} 
        & Candidate trajectories ($K$) & 32 \\
        & Route completion threshold ($\eta_{rc}$) & 0.85 \\
        & Marginal preference threshold ($\epsilon$) & 0.02 \\
        & DPO temperature parameter ($\beta$) & 0.05 \\
        \midrule
        \multirow{2}{*}{Capability Prediction} 
        & Standardized probe states ($M$) & 149 \\
        & Historical state sequence length ($L$) & 20 \\
        \midrule
        \multirow{5}{*}{Curriculum Sampling} 
        & Maximum training horizon ($T_{max}$) & 1 $ \times 10^6$ \\
        & Prediction update interval ($N_{eval}$) & 2.5 $ \times 10^4$ \\
        & Base scenario capacity limit ($N_{max}$) & 5 \\
        & Boundary trigger threshold ($\rho_{min}$) & 0.1 \\
        & Adversarial scaling factor ($\kappa$) & 2 \\
        \bottomrule
    \end{tabular}
\end{table}

We benchmark the proposed framework against established baseline methods in both scenario generation and sampling domains. In terms of scenario generation, the proposed resolvable generator is compared against a deterministic log replay method and the CAT \citep{zhang2023cat} framework. For scenario sampling, we evaluate against a random sampling baseline, a scenario-based static sampling strategy based on predefined scenario complexity and risk metrics, and a history-based sampling strategy that relies on the historical performance of the ego agent in specific scenarios. The overall performance of the trained driving policies is evaluated using core metrics, including crash rate, route completion rate, average reward, average cost, and convergence speed.

\subsection{Verification of Scenario Solvability}

\begin{figure}[h]
    \centering
    \includegraphics[width=0.95\linewidth]{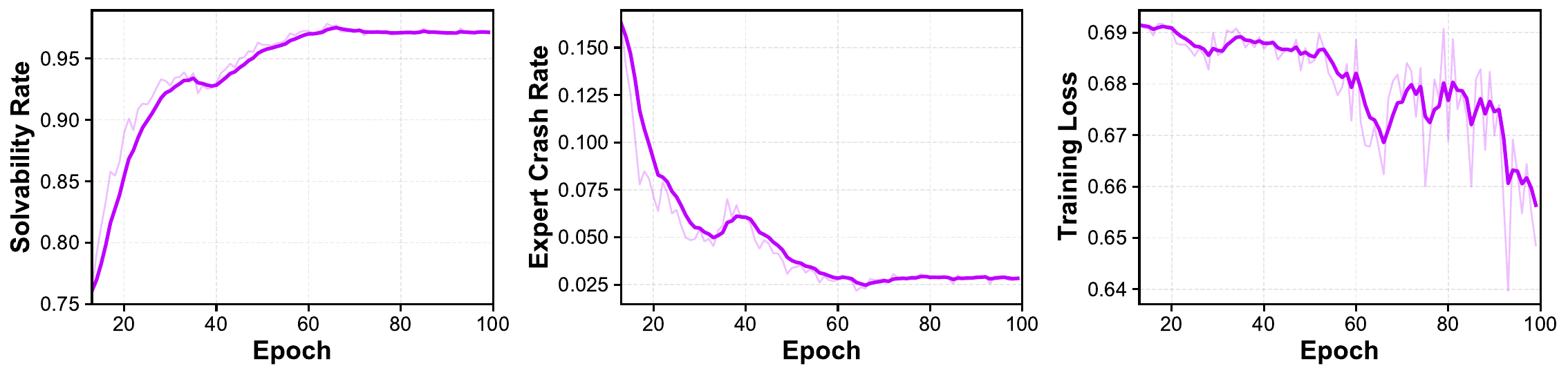}
    \caption{Evolution of key performance metrics during the DPO fine-tuning process.}
    \label{fig:dpo_training_curves}
\end{figure}

The primary objective of our scenario generation module is to ensure that the generated adversarial scenarios present critical but resolvable challenges for the autonomous driving policy. To validate the effectiveness of the fine-tuned trajectory prediction model, we first analyze the training dynamics of the preference optimization process. As illustrated in Figure \ref{fig:dpo_training_curves}, the solvability rate steadily increases and converges near a perfect score, while the expert crash rate experiences a significant and rapid decline. Here, the expert crash rate refers to the collision rate of the privileged rule-based expert when it is rolled out as the ego controller in the generated scenarios. Concurrently, the training loss decreases steadily, indicating that the trajectory prediction model successfully aligns with the preferences of the rule-based expert. The increasing solvability rate indicates that the generated trajectories for the adversarial vehicles leave at least one solution for the ego agent.

A detailed quantitative comparison between the pre-trained and fine-tuned models further validates the effectiveness of our approach, as presented in Table \ref{tab:model_comparison}. The most prominent improvement lies in  the absolute number of unsolvable scenarios, which drops from over six hundred in the pre-trained model to merely twenty in our fine-tuned model. Consequently, the overall solvability rate increases to nearly one hundred percent. Importantly, this enhancement in solvability does not compromise the behavioral diversity of the adversarial vehicles. The diversity metrics, including the Average Pairwise Distance (APD), the Final Pairwise Distance (FPD), and the trajectory variance, remain highly consistent between the two models. The APD measures the mean spatial deviation among the generated trajectories over the entire time horizon, whereas the FPD evaluates the dispersion of the trajectory endpoints. The preservation of these metrics indicates that the fine-tuning process refines the physical feasibility of the trajectories without sacrificing the overall diversity of the adversarial behaviors.

\begin{table}[h]
    \centering
    \small
    \caption{Quantitative comparison of scenario generation performance between Pre-trained and Fine-tuned models.}
    \label{tab:model_comparison}
    \setlength{\tabcolsep}{4pt} 
    \begin{tabular}{lcccccc}
        \toprule
        \multirow{2}{*}{Model} & \multirow{2}{*}{\makecell{Unsolvable Scenarios\\ (Count) }$\downarrow$} & \multirow{2}{*}{\makecell{Solvability Rate\\ (\%) }$\uparrow$} & \multirow{2}{*}{\makecell{Average TTC\\ (s)}$\downarrow$} & \multicolumn{3}{c}{Diversity Metrics} \\
        \cmidrule(lr){5-7}
        & & & & APD (m) $\uparrow$& FPD (m) $\uparrow$& Variance $\uparrow$\\
        \midrule
        Pre-trained & 606 & 96.05 & 0.582 & 2.771 & 8.369 & 2.696 \\
        Fine-tuned   & \textbf{20} & \textbf{99.87} & 0.577 & 2.709 & 8.284 & 2.779 \\
        \bottomrule
    \end{tabular}
\end{table}

\begin{figure}[hb!]
    \centering
    \includegraphics[width=0.5\linewidth]{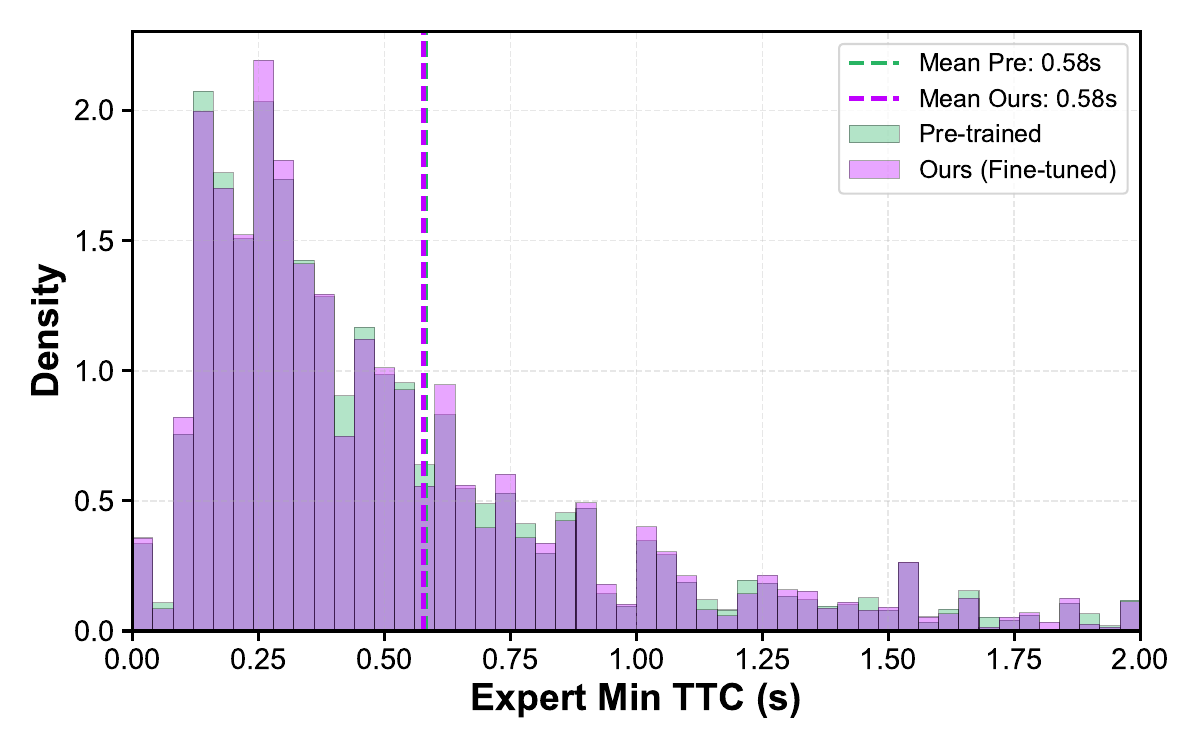}
    \caption{Density distribution of the minimum TTC for scenarios generated before and after fine-tuning.}
    \label{fig:dpo_ttc_distribution_comparison}
\end{figure}

Furthermore, it is essential to ensure that the newly generated solvable scenarios remain sufficiently challenging to drive the improvement of the autonomous driving policy. We utilize TTC to quantify the risk level of the generated scenarios. A comparison of the probability density distributions of the minimum TTC 
is depicted in Figure \ref{fig:dpo_ttc_distribution_comparison}. The average minimum TTC is remarkably stable at approximately 0.58 seconds for both distributions. The distribution profile of the fine-tuned model closely aligns with that of the pre-trained model, which confirms that the optimization process specifically eliminates fundamentally unsolvable corner cases while successfully preserving the overall adversarial value of the training environment.

\begin{figure}[hb!]
    \centering
    \includegraphics[width=0.75\linewidth]{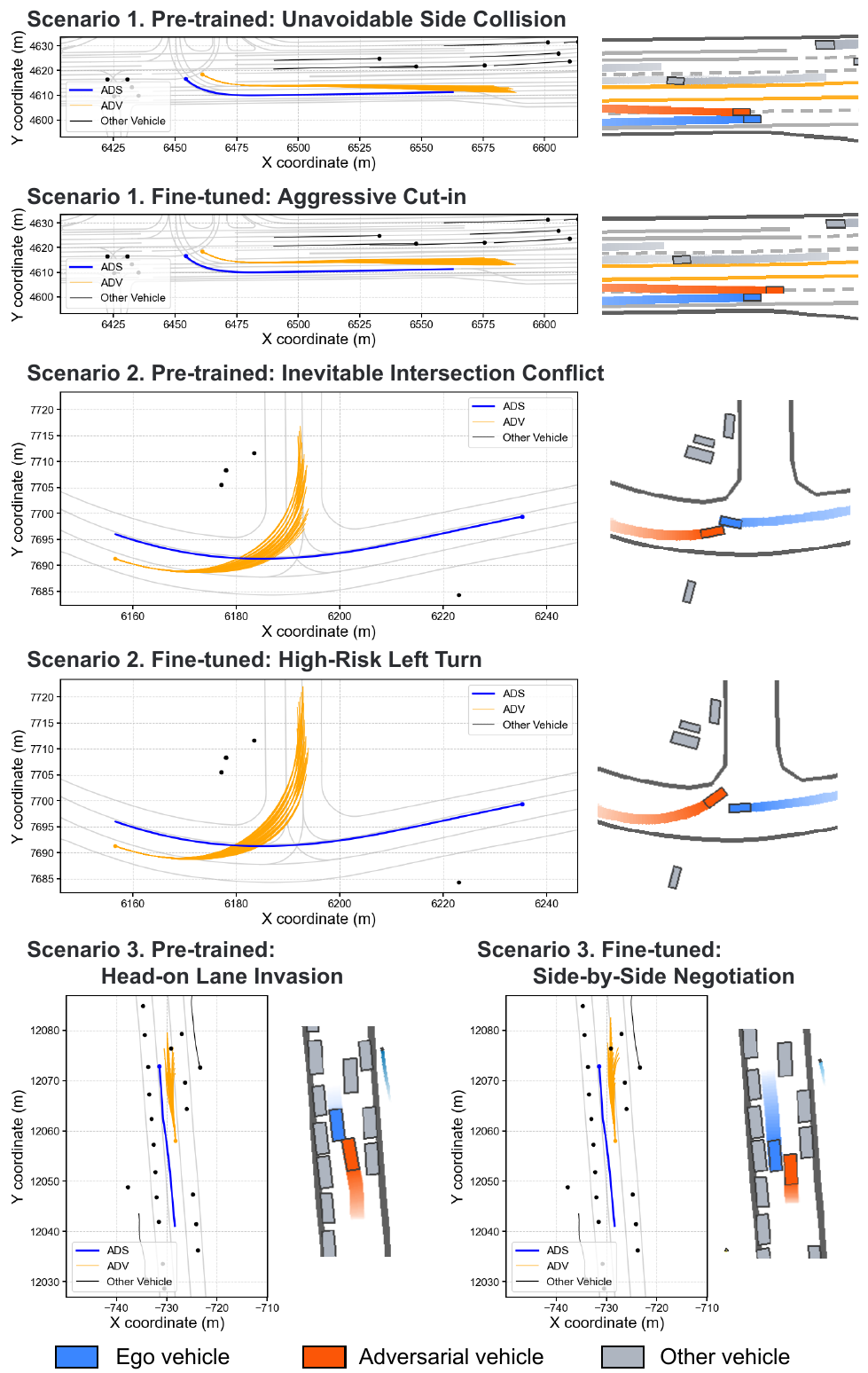}
    \caption{Qualitative comparison of adversarial scenarios generated by the pre-trained model and our fine-tuned model.}
    \label{fig:solve_case}
\end{figure}

The qualitative advantages of our approach are further demonstrated through specific traffic conflict cases. As shown in Figure \ref{fig:solve_case}, the pre-trained model frequently produces physically unreasonable situations. For instance, it generates unavoidable side collisions where the background vehicle directly crashes into the ego vehicle from an adjacent lane. Such extreme situations force the ego vehicle into inevitable crashes. In contrast, the fine-tuned model adjusts these behaviors into aggressive cut-ins, high-risk left turns during intersection conflicts, and tight side-by-side negotiations. These refined scenarios demand precise control and advanced anticipation from the ego vehicle, thereby fostering a more robust and capable autonomous driving policy.

\subsection{Latent Representations of the Behavioral Fingerprint}
\begin{figure}
    \centering
    \includegraphics[width=0.48\linewidth]{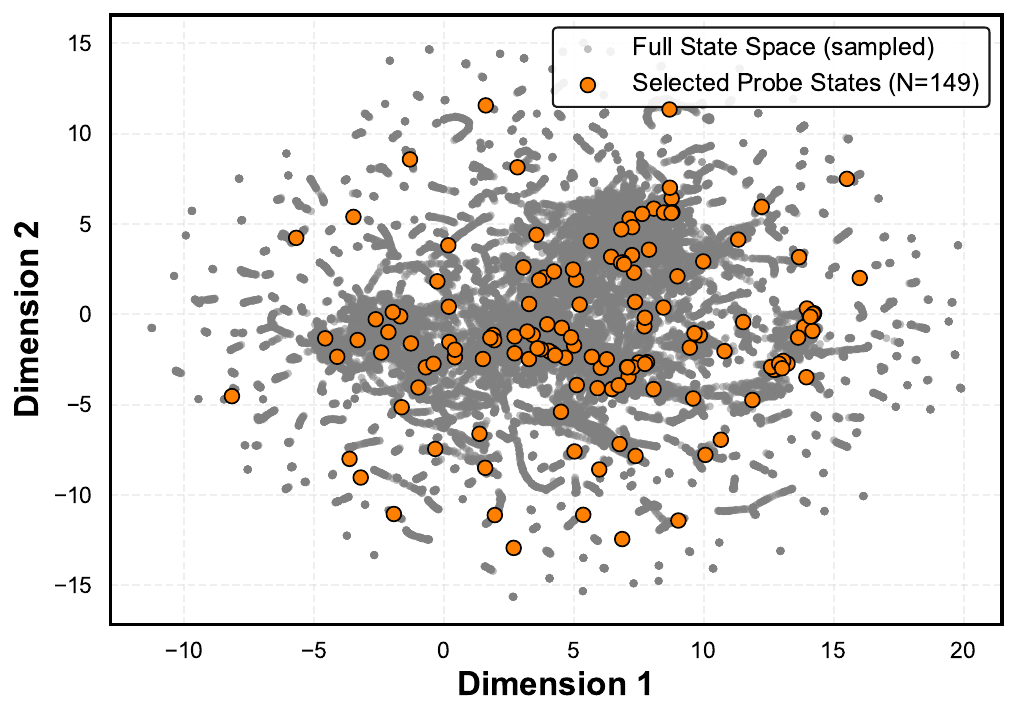}
    \caption{UMAP Visualization of the Selected Probe States. The scatter plot illustrates the distribution of the 149 selected probe states relative to the full sampled state space.}
    \label{fig:fingerprint_probe_state_distribution}
\end{figure}

To effectively sample scenarios based on the dynamic capabilities of the autonomous driving policy, we propose the behavioral fingerprint mechanism designed to capture the intrinsic driving proficiency of the policy. The foundation of this mechanism relies on a meticulously selected set of probe states. Figure \ref{fig:fingerprint_probe_state_distribution} visualizes the uniform manifold approximation and projection of the full state space alongside the selected probe states. The selected 149 probe states are thoroughly distributed across the entire sampled state space, ensuring that the resulting behavioral fingerprint comprehensively captures the driving characteristics of the algorithm under diverse and critical conditions.

\begin{figure}[!htbp]
    \centering
    \begin{subfigure}{0.473\linewidth}
        \centering
        \includegraphics[width=\linewidth]{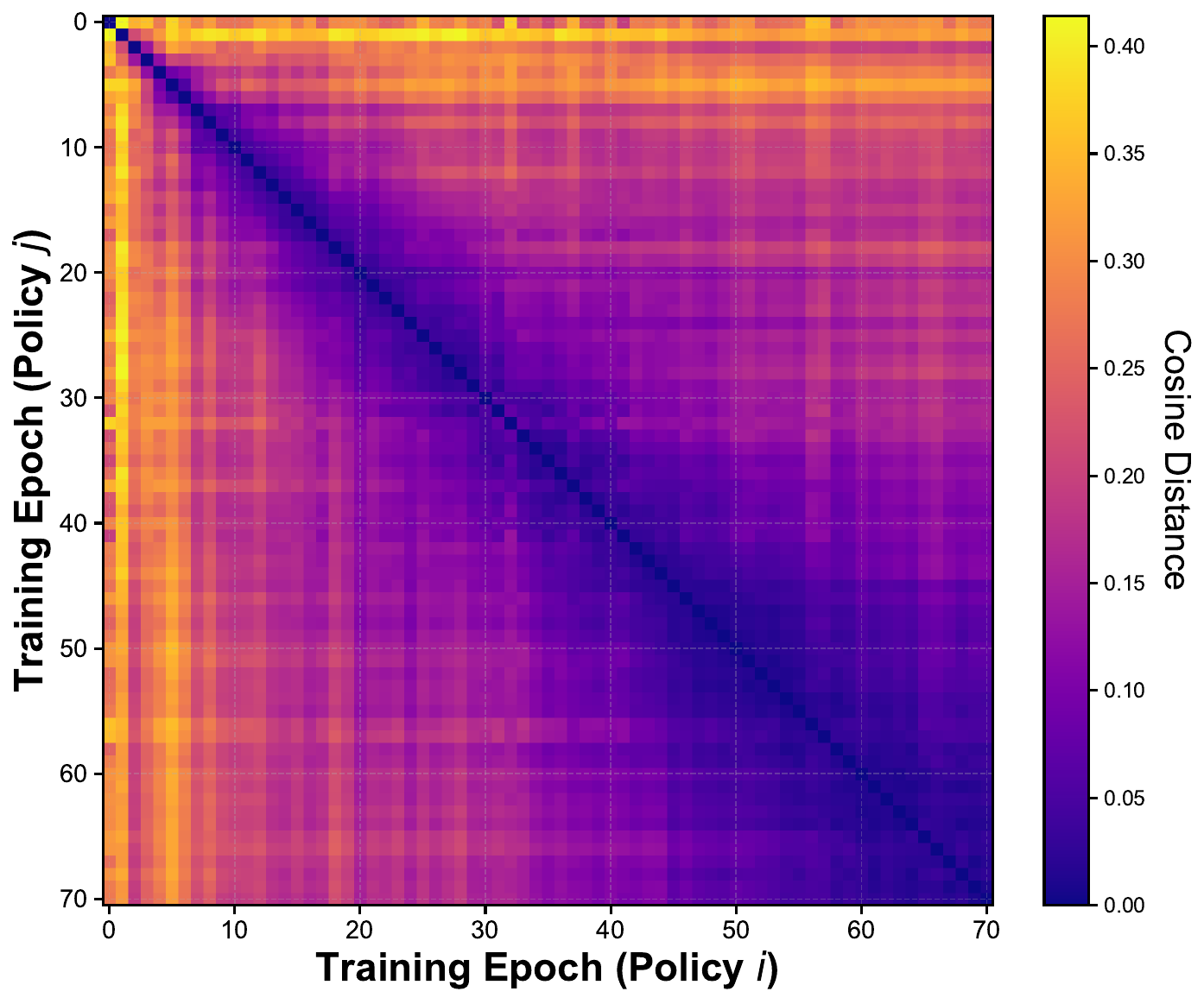}
        \caption{Cosine distance matrix.}
        \label{fig:fingerprint_cosine_heatmap}
    \end{subfigure}
    \hfill
    \begin{subfigure}{0.48\linewidth}
        \centering
        \includegraphics[width=\linewidth]{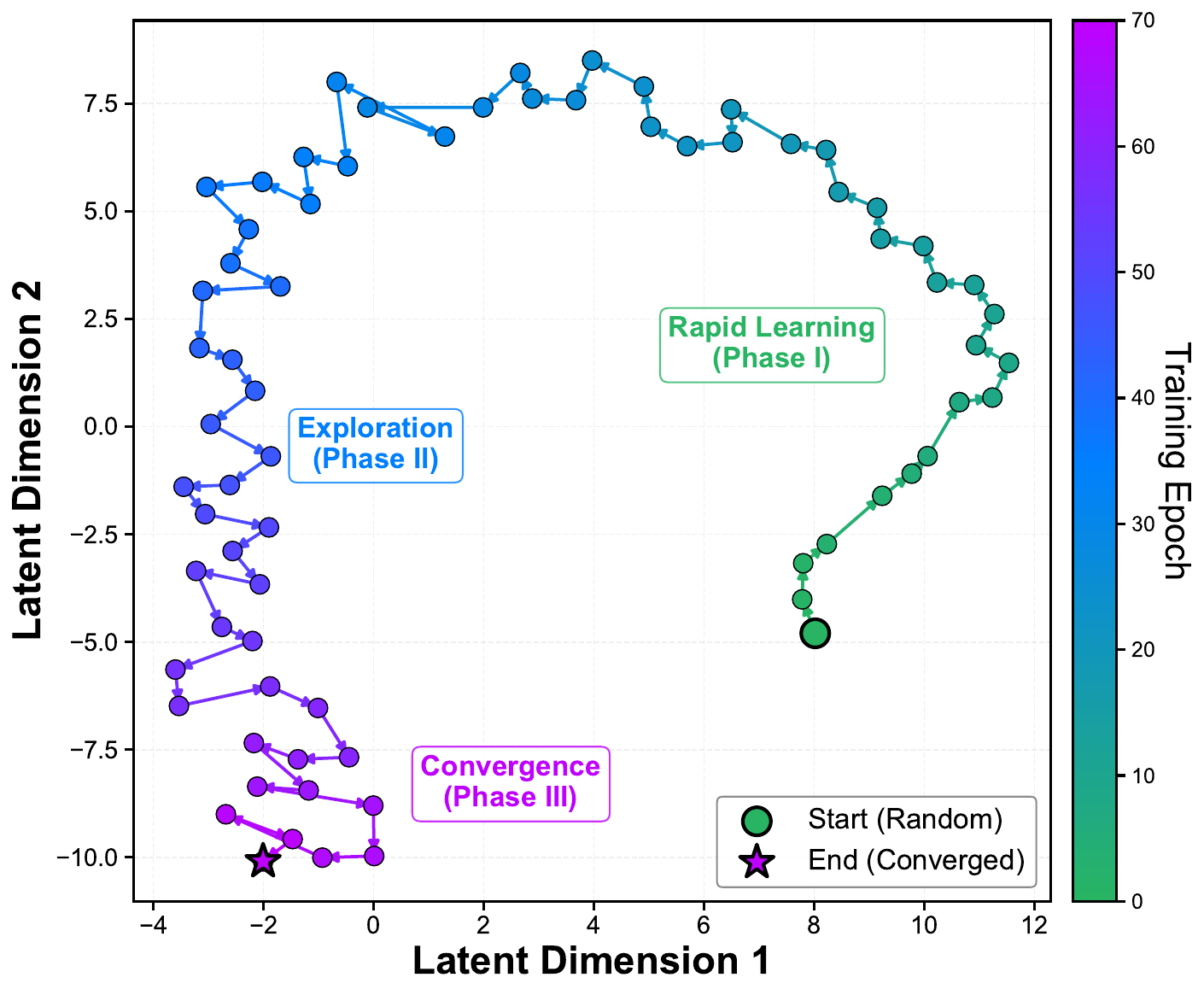}
        \caption{Evolution of Behavioral Fingerprints.}
        \label{fig:fingerprint_trajectory}
    \end{subfigure}
    \caption{Evolutionary Analysis of Behavioral Fingerprints. The cosine distance matrix highlights the similarity between policies at different training epochs, showing clear convergence. The t-SNE trajectory visualizes the continuous learning dynamics, distinctly capturing the rapid learning, exploration, and convergence phases.}
    \label{fig:fingerprint}
\end{figure}

The effectiveness of the behavioral fingerprint is first evidenced by its discriminative power across different training stages. We collect the fingerprints of the policy at regular intervals throughout the training process and compute their pairwise cosine distances, as illustrated in Figure \ref{fig:fingerprint_cosine_heatmap}. For two policy checkpoints $\pi_{\theta_i}$ and $\pi_{\theta_j}$ with behavioral fingerprints $\mathbf{f}_{\theta_i}$ and $\mathbf{f}_{\theta_j}$, the matrix entry is computed as $d_{ij}=1-\frac{\mathbf{f}_{\theta_i}^\top \mathbf{f}_{\theta_j}}{|\mathbf{f}_{\theta_i}||\mathbf{f}_{\theta_j}|}$. The matrix reveals distinct visual patterns characterized by dark regions indicating high similarity and bright regions indicating significant divergence. Specifically, the fingerprints from the early training epochs exhibit immense distances from those in the later epochs, proving that the fingerprint can easily differentiate an untrained policy from a mature policy. Furthermore, the large contiguous dark block in the lower right corner of the matrix indicates that the distances between epochs become negligible during the late stages of training. This phenomenon confirms that the behavioral fingerprint successfully captures the convergence properties and the stabilization of the algorithmic policy.

To further understand the evolutionary dynamics embedded within the behavioral fingerprints, we project the high-dimensional fingerprint vectors into a two-dimensional space using the t-distributed stochastic neighbor embedding (t-SNE) technique. As depicted in Figure \ref{fig:fingerprint_trajectory}, the temporal sequence of the fingerprints forms a continuous and smooth trajectory rather than a random scattering of points. This trajectory clearly delineates three distinct phases of the training process. The initial rapid descent phase shows the policy quickly departing from its random initialization. This is followed by an exploration phase characterized by oscillatory and sweeping movements in the latent space, reflecting the trial-and-error nature of reinforcement learning. Finally, the trajectory enters a convergence phase where the data points become highly clustered, signifying that the driving behavior has stabilized. This structured evolution proves that the fingerprint maps to a physically meaningful feature space that reflects the underlying learning mechanism.

\begin{figure}
    \centering
    \includegraphics[width=0.46\linewidth]{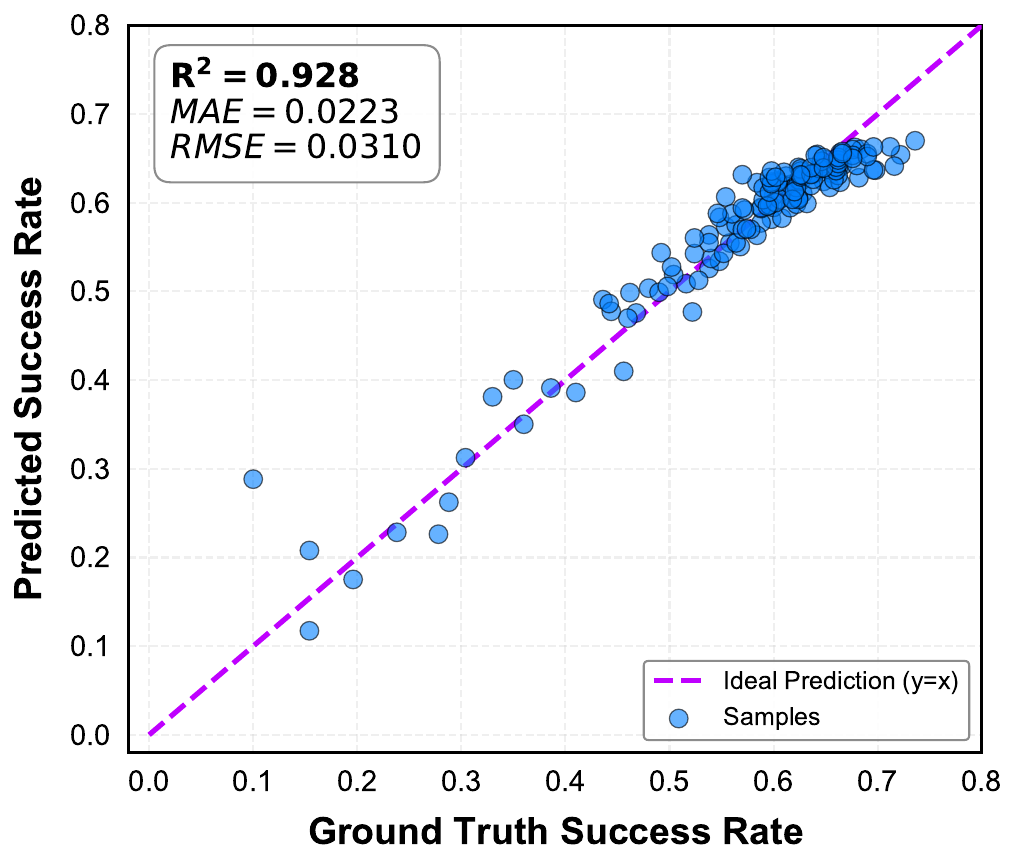}
    \caption{Correlation between predicted and ground truth success rates.}
    \label{fig:fingerprint_prediction_scatter}
\end{figure}

The ultimate validation of the behavioral fingerprint lies in its ability to directly predict the actual driving capabilities of the policy. We trained a multilayer perceptron model exclusively utilizing the fingerprint to forecast the average success rate of various policies across a diverse scenario library. Utilizing a dataset of one hundred and forty six distinct policies and employing a five-fold cross-validation method, the prediction model achieves an impressive Mean Absolute Error of 0.0223. The high correlation depicted in the prediction scatter plot in Figure \ref{fig:fingerprint_prediction_scatter} confirms that the fingerprint inherently contains rich semantic information. These results prove that the behavioral fingerprint is a highly effective feature representation, capable of guiding the dynamic scenario sampling process by accurately reflecting the current competence of the autonomous driving algorithm.

\subsection{Evaluation of Capability Prediction}
To comprehensively assess the performance of the capability predictor, we employ a diverse set of evaluation metrics to capture both the overall classification accuracy and the specific requirements of the downstream curriculum sampling task. Best Accuracy and the Area Under the Curve (AUC) serve as fundamental indicators of the overall correctness and the general ability of the model to rank the predicted probabilities across various classification thresholds. 
Furthermore, we compute the average precision for both failure and positive outcomes to assess the robustness of the predictor.
Most critically, the ultimate goal of this predictor is to identify challenging scenarios to guide the curriculum sampling process. Consequently, accurately predicting impending failures is important. To reflect this specific requirement, we introduce the Precision at the top ten percent of negative predictions, which measures the reliability of the model when it is most confident that an agent will fail. A high value in this specific metric indicates that the scenarios prioritized by our sampling strategy are genuinely challenging for the current policy. 

\begin{figure}[ht]
    \centering
    \begin{subfigure}{0.48\linewidth}
        \centering
        \includegraphics[width=\linewidth]{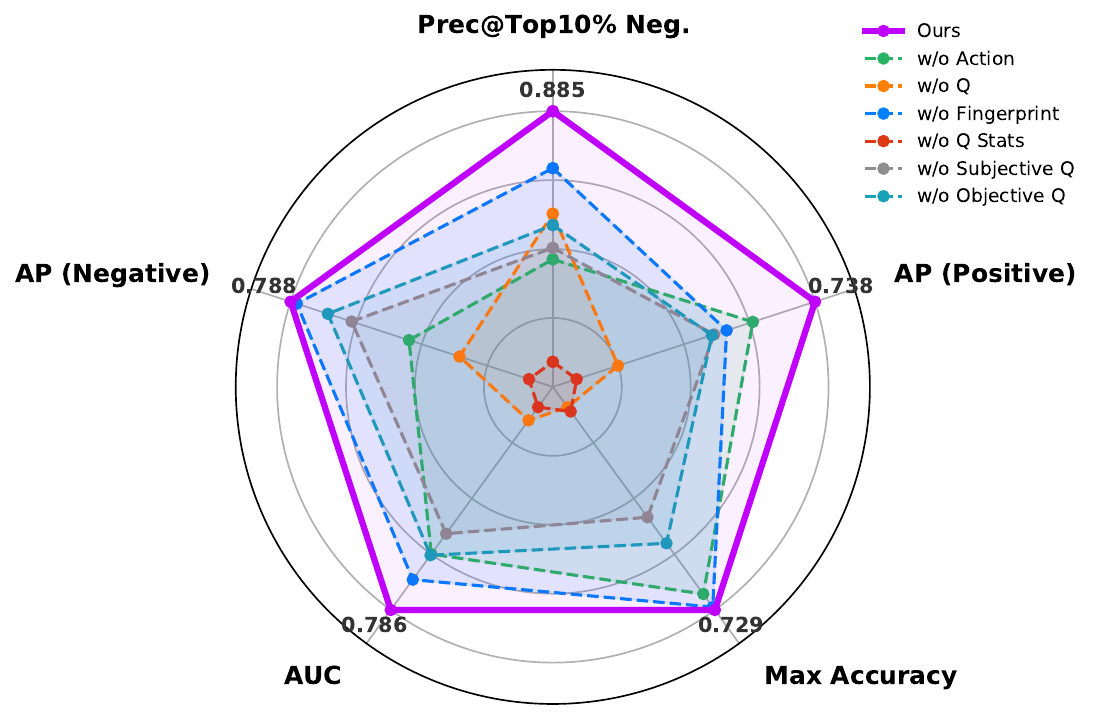}
        \caption{Feature ablation variants.}
        \label{fig:predictor_radar}
    \end{subfigure}
    \hfill
    \begin{subfigure}{0.48\linewidth}
        \centering
        \includegraphics[width=\linewidth]{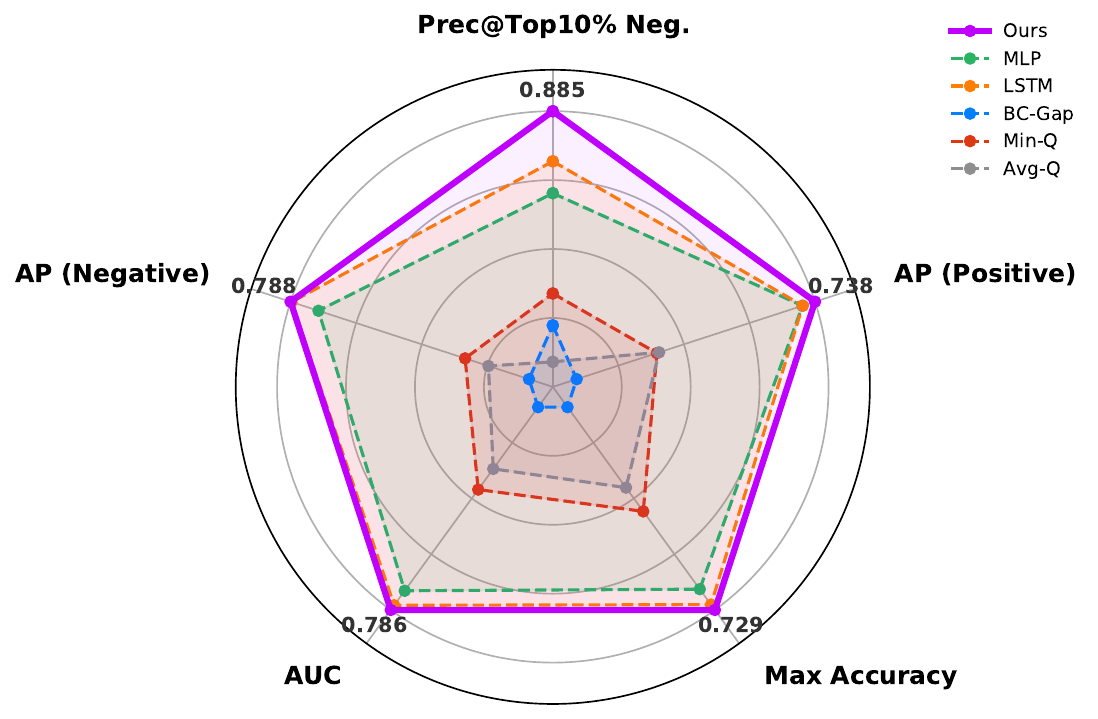}
        \caption{Comparison with baseline methods.}
        \label{fig:predictor_radar_other_models}
    \end{subfigure}
    \caption{Radar charts comparing the comprehensive performance of different predictor architectures.}
    \label{fig:fingerprint_2}
\end{figure}

Building upon these established metrics, we conduct a comprehensive comparison of our proposed model against several heuristic baselines and structural variants. As detailed in Table \ref{tab:ablation_study} and illustrated through the radar charts in Figure \ref{fig:fingerprint_2}, the multi-modal capability predictor significantly outperforms heuristic approaches such as behavior cloning gap and static average or minimum Q-value estimations. 
Furthermore, comparing the structural variants reveals that while recurrent architectures like Long Short-Term Memory networks (LSTM) offer improvements over simple Multi-Layer Perceptrons (MLP), they still fall short of our proposed cross-attention fusion mechanism.
Most notably, the full model attains an outstanding precision of 0.885 in the top ten percent of negative predictions, fundamentally validating its capacity to reliably guide the curriculum sampler toward the most valuable training scenarios tailored to the evolving weaknesses of the ego vehicle.

\begin{figure}[!h]
    \centering
    \includegraphics[width=0.95\linewidth]{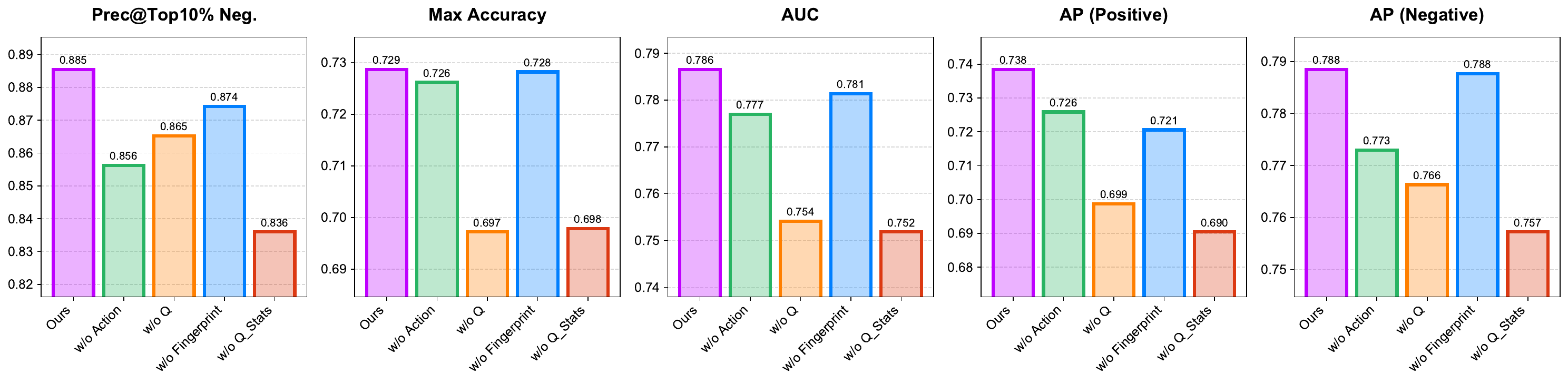}
    \caption{Ablation Study on Capability Predictor Features.}
    \label{fig:predictor_ablation}
\end{figure}

To validate the architectural design and the contribution of individual feature components, we conduct feature ablation studies. As illustrated in Figure \ref{fig:predictor_ablation} and Table \ref{tab:ablation_study}, the removal of any designed multimodal feature leads to a distinct degradation in predictive accuracy. 
The elimination of the behavior fingerprint results in a notable drop in the ability to precisely identify failure cases and significantly decreases the average precision for positive cases, proving that the global representation derived from probe states is essential. 
Similarly, omitting localized scenario contexts, such as the actual action sequences, the Q-value sequence or the Q-statistics, compromises the precision of the predictor. The ablation results also indicate that relying exclusively on either subjective expert evaluations or objective self-evaluations is less precise than the combined approach.
These findings confirm that the localized scenario context and the global policy characteristics are highly complementary, and their fusion is indispensable for precise capability estimation.

\begin{table*}[htbp]
\centering
\small
\caption{Performance comparison of the capability predictor against heuristic baselines and ablation variants. The evaluation metrics include best accuracy, area under the curve, average precision for negative and positive cases, and the precision for identifying the top ten percent most challenging negative scenarios. \textbf{BC-Gap}: Behavior Cloning Gap (Action Similarity); \textbf{Min-Q}: Conservative Q-Estimate; \textbf{Avg-Q}: Average Q-Estimate. The best results are highlighted in \textbf{bold}.}
\label{tab:ablation_study}
\renewcommand{\arraystretch}{1} 
\begin{tabular}{lccccc}
\toprule
\textbf{Model Variant} & \textbf{Best Acc.} & \textbf{AUC} & \textbf{AP (Negative)} & \textbf{AP (Positive)} & \textbf{Prec@Top10\% Neg.} \\
\midrule
\multicolumn{6}{l}{\textit{Heuristic Baselines}} \\
BC-Gap & 0.575 & 0.597 & 0.651 & 0.516 & 0.780 \\
Avg-Q & 0.636 & 0.655 & 0.674 & 0.593 & 0.762 \\
Min-Q & 0.654 & 0.674 & 0.688 & 0.591 & 0.796 \\
\midrule
\multicolumn{6}{l}{\textit{Structural Variant}} \\
MLP Backbone & 0.713 & 0.769 & 0.772 & 0.727 & 0.845 \\
LSTM Backbone & 0.725 & 0.782& \textbf{0.788} & 0.727 & 0.861 \\
\midrule
\multicolumn{6}{l}{\textit{Feature Ablation Study}} \\
w/o Action Sequence & 0.726 & 0.777 & 0.773 & 0.726 & 0.856 \\
w/o Q-Value Sequence & 0.697 & 0.754 & 0.766 & 0.699 & 0.865 \\
w/o Behavior Fingerprint & 0.728 & 0.781 & \textbf{0.788} & 0.721 & 0.874 \\
w/o Q-Statistics & 0.698 & 0.752 & 0.757 & 0.690 & 0.836 \\
\midrule
\multicolumn{6}{l}{\textit{Q-Value Component Analysis}} \\
w/o Subjective Q (Only Expert) & 0.714 & 0.773 & 0.780 & 0.718 & 0.858 \\
w/o Objective Q (Only Self) & 0.718 & 0.777 & 0.784 & 0.718 & 0.863 \\
\midrule
\textbf{Ours (Full Model)} & \textbf{0.729} & \textbf{0.786} & \textbf{0.788} & \textbf{0.738} & \textbf{0.885} \\
\bottomrule
\end{tabular}
\end{table*}

\subsection{Closed-loop Adversarial Training Results}
\begin{figure}[h!]
    \centering
    \includegraphics[width=0.95\linewidth]{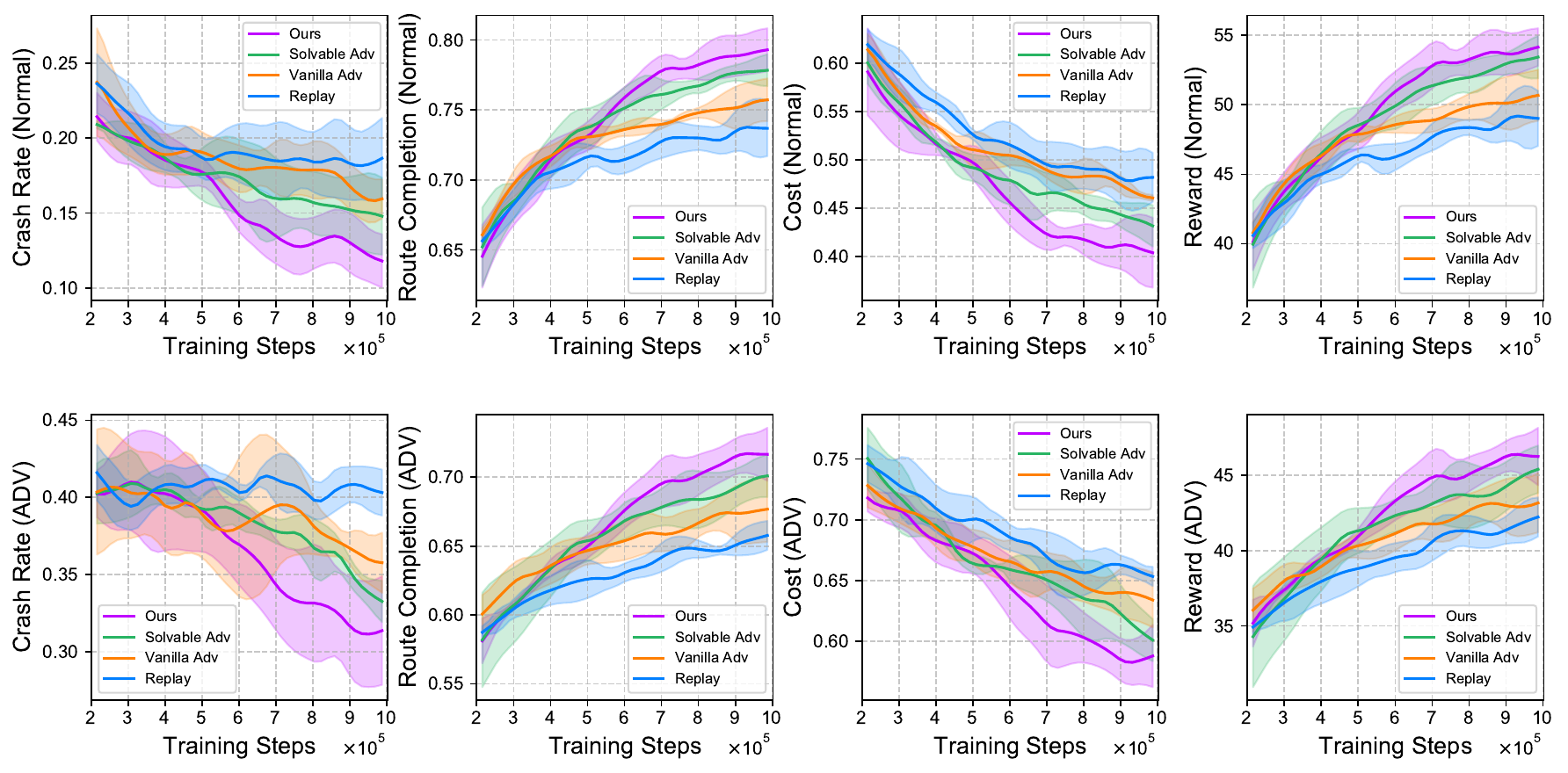}
    \caption{Learning curves of the autonomous driving policy under different scenario generation methods. The solid lines represent the mean performance across multiple random seeds, and the shaded areas indicate the standard deviation. The proposed framework achieves faster convergence and higher performance in both normal and adversarial environments.}
    \label{fig:RL_train_comparison}
\end{figure}

\begin{figure}[h!]
    \centering
    \includegraphics[width=0.95\linewidth]{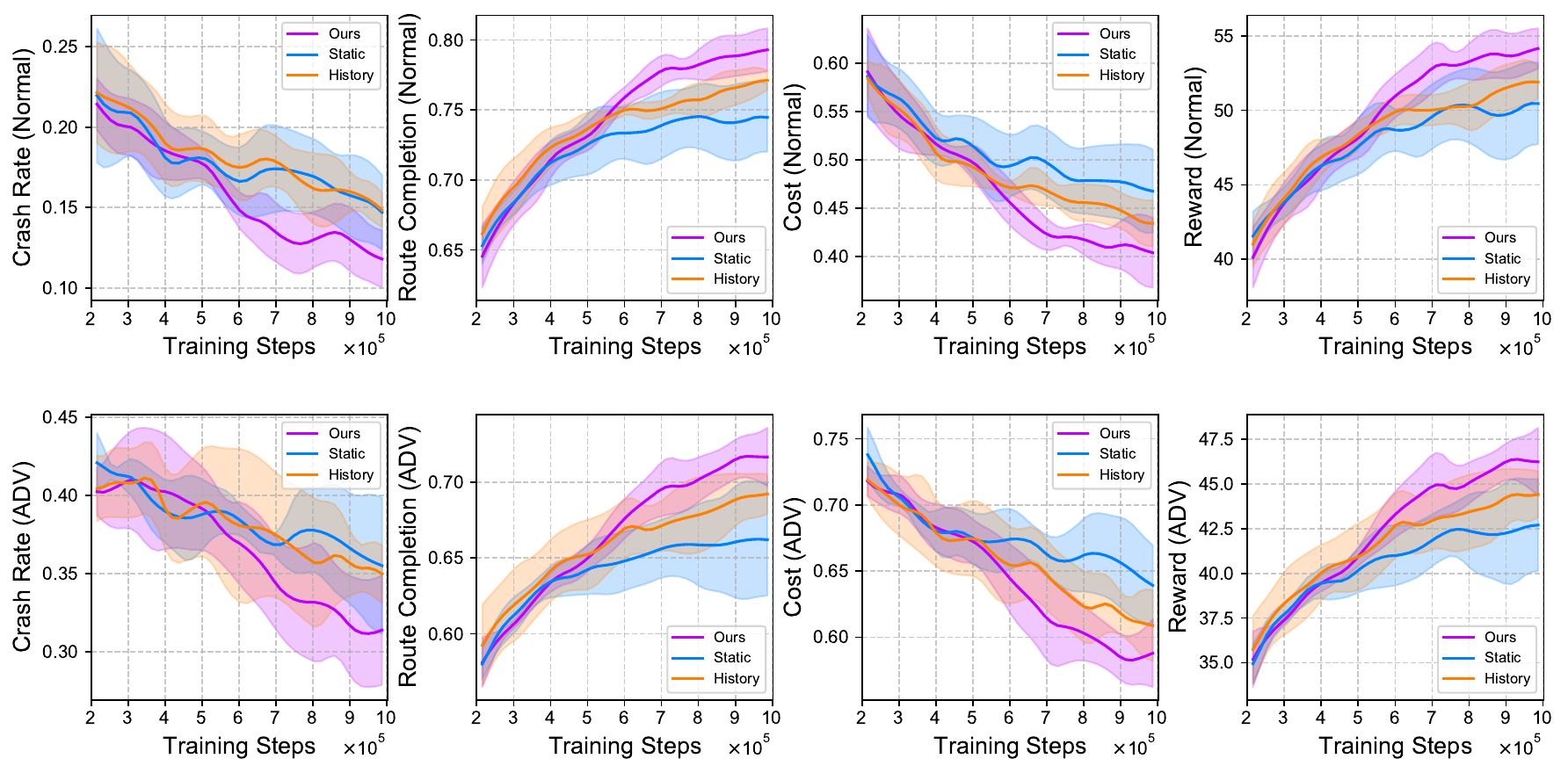}
    \caption{Impact of different curriculum sampling strategies on reinforcement learning training. All methods utilize the proposed solvable adversarial scenarios but differ in how scenarios are selected. The curves demonstrate that dynamic capability-aligned sampling accelerates convergence and achieves superior final performance compared to static and history-based sampling methods.}
    \label{fig:RL_train_sample_methods_comparison}
\end{figure}

The ultimate objective of our proposed framework is to enhance the training efficacy and the performance of the autonomous driving policy through high-quality scenario generation and dynamic curriculum sampling. We systematically evaluate the reinforcement learning training outcomes under both normal and adversarial environmental configurations. Figure \ref{fig:RL_train_comparison}, Figure \ref{fig:RL_train_sample_methods_comparison}, Table \ref{tab:combined_performance}, and Table \ref{tab:combined_performance_adv} present a comprehensive comparison between our proposed framework and other baselines. 

The advantages of our framework are fundamentally rooted in the introduction of the resolvable adversarial scenario generator. We first evaluate the direct impact of the solvable scenario generation mechanism (Solvable Adv) by comparing it against the Replay and the unconstrained adversarial training baseline (CAT, denoted as Vanilla Adv) under a standard random sampling strategy. Relying solely on replay scenarios results in the lowest overall performance, evidenced by a crash rate of 18.65 \% in normal scenarios and 40.31 \% in adversarial settings. The unconstrained adversarial baseline reduces the crash rates to 15.95 \% and 35.77 \%, respectively, while simultaneously increasing the route completion rates. Our resolvable adversarial generator achieves further substantial improvements. The policy trained with solvable scenarios achieves a crash rate of 14.79\% and a route completion rate of 77.84\% in normal conditions. In the more challenging adversarial environments, the resolvable generator reduces the crash rate to 33.27\% and elevates the route completion to 70.10\%. These quantitative results confirm that generating solvable adversarial scenarios directly enhances the final performance metrics compared to unconstrained adversarial generation.

Building upon the resolvable scenario generator, we evaluate the effectiveness of the dynamic capability predictor in guiding the curriculum sampling distribution. The experimental results demonstrate that predicting the success probability at the current training step yields superior policy performance compared to static or history-based sampling strategies. The static sampling method yields a crash rate of 14.70 \% and a route completion rate of 74.47 \% in normal scenarios. The historical performance-based sampling achieves a crash rate of 14.85 \% and a route completion rate of 77.13 \%. In comparison, our proposed dynamic curriculum sampling method achieves the best performance across all metrics. It achieves the lowest overall crash rates of 11.80\% in normal scenarios and 31.37\% in adversarial scenarios. Concurrently, the route completion rates reach peak values of 79.30\% and 71.66\% across the two testing environments, validating the functional advantage of capability prediction for scenario selection.

Beyond final performance metrics, the integration of the resolvable generator and the dynamic capability predictor fundamentally accelerates the convergence speed of the reinforcement learning process. We quantify training efficiency by recording the number of simulation steps required to reach specific performance thresholds, specifically a crash rate below 16\% and a route completion rate above 75\% in normal scenarios. The standard replay methods fail to achieve the safety threshold within the maximum allocated training steps. The vanilla adversarial method requires over 950,000 steps to reach the safety target in both environments. By employing our complete proposed framework, the required training duration to reach the crash rate threshold drastically drops to 564,274 steps, translating to a convergence speedup of 1.68 times and a 40.6\% savings in training steps compared to the vanilla adversarial baseline. Similarly, the necessary steps to reach the route completion target are reduced to 582,638 steps, saving 32.1\% of the required steps over the baseline. 

Overall, compared to the adversarial training baseline, our fully integrated framework consumed only 59 \% of the training steps to reach the established performance thresholds and ultimately improved the final safety performance by delivering a 26 \% relative reduction in the crash rate alongside persistent enhancements in route completion.


\begin{table}[h]
    \centering
    \small
    \caption{Performance evaluation of the trained policies in \textbf{Normal} scenarios. The upper block compares different scenario generation methods (with random sampling), while the lower block compares different curriculum sampling strategies (with our solvable generator). The ``Steps Saved'' indicates the reduction in training steps required to reach the crash rate threshold relative to the Vanilla Adv baseline.}
    \label{tab:combined_performance}
    \renewcommand{\arraystretch}{1.1} 
    \setlength{\tabcolsep}{3pt} 
    \begin{tabular}{lccccccc}
        \toprule
        \multirow{3}{*}{Method} & \multicolumn{4}{c}{Performance Metrics (Mean $\pm$ Std)} & \multicolumn{3}{c}{Training Efficiency (Steps)} \\
        \cmidrule(lr){2-5} \cmidrule(lr){6-8}
        & Crash Rate $\downarrow$ & Route Comp. $\uparrow$ & Cost $\downarrow$ & Reward $\uparrow$ & \makecell{Crash Rate \\ ($< 0.16$)} $\downarrow$ & \makecell{Route Comp. \\ ($> 0.75$)} $\downarrow$ & \makecell{Steps \\ Saved}$\uparrow$ \\
        \midrule
        \multicolumn{4}{l}{\textit{Comparison of Generators}} & & & \\
        Replay        & 18.65\% $\pm$ 2.70\%           & 73.69\% $\pm$ 1.99\%           & 0.482 $\pm$ 0.026          & 49.01 $\pm$ 1.99          & $>$ Max Steps    & $>$ Max Steps   & -- \\
        Vanilla Adv      & 15.95\% $\pm$ 1.30\%           & 75.73\% $\pm$ 1.53\%           & 0.461 $\pm$ 0.003          & 50.68 $\pm$ 1.84          & 950,002          & 858,170   & Baseline       \\
        Solvable Adv      & 14.79\% $\pm$ 2.57\%           & 77.84\% $\pm$ 1.14\%           & 0.432 $\pm$ 0.021          & 53.43 $\pm$ 1.53          & 692,842          & 601,022        & 27.1\%  \\
        \midrule
        \multicolumn{4}{l}{\textit{Comparison of Samplers}} & & & \\
        Static        & 14.70\% $\pm$ 2.32\%           & 74.47\% $\pm$ 2.44\%           & 0.468 $\pm$ 0.043          & 50.45 $\pm$ 2.73          & 876,497    & $>$ Max Steps & 7.7\%   \\
        History       & 14.85\% $\pm$ 1.07\%           & 77.13\% $\pm$ 0.69\%           & 0.434 $\pm$ 0.024          & 51.90 $\pm$ 1.07          & 913,249          & 601,062    & 3.9\%      \\
        \midrule
        \textbf{Ours} & \textbf{11.80\% $\pm$ 1.78\% } & \textbf{79.30\% $\pm$ 1.54\% } & \textbf{0.404 $\pm$ 0.036} & \textbf{54.15 $\pm$ 1.36} & \textbf{564,274} & \textbf{582,638} & \textbf{40.6\%} \\
        \bottomrule
    \end{tabular}
\end{table}


\begin{table}[h]
    \centering
    \small
    \caption{Performance evaluation of the trained policies in \textbf{Adversarial} scenarios. The upper block compares different scenario generation methods (with random sampling), while the lower block compares different curriculum sampling strategies (with our solvable generator).}
    \label{tab:combined_performance_adv}
    \renewcommand{\arraystretch}{1.1} 
    \setlength{\tabcolsep}{4pt} 
    \begin{tabular}{lccccccc}
        \toprule
        \multirow{3}{*}{Method} & \multicolumn{4}{c}{Performance Metrics (Mean $\pm$ Std)} & \multicolumn{3}{c}{Training Efficiency (Steps)} \\
        \cmidrule(lr){2-5} \cmidrule(lr){6-8}
        & Crash Rate $\downarrow$ & Route Comp. $\uparrow$ & Cost $\downarrow$ & Reward $\uparrow$ & \makecell{Crash Rate \\ ($< 0.36$)} $\downarrow$ & \makecell{Route Comp. \\ ($> 0.67$)} $\downarrow$ & \makecell{Steps \\ Saved}$\uparrow$ \\
        \midrule
        \multicolumn{4}{l}{\textit{Comparison of Generators}} & & & \\
        Replay        & 40.31\%  $\pm$ 1.49\%           & 65.77\%  $\pm$ 1.05\%          & 0.653 $\pm$ 0.008          & 42.23 $\pm$ 1.30          & $>$ Max Steps    & $>$ Max Steps & --    \\
        Vanilla Adv      & 35.77\%  $\pm$ 1.95\%           & 67.68\%  $\pm$ 2.39\%           & 0.634 $\pm$ 0.023         & 43.16 $\pm$ 1.96          & 950,002          & 839,804    & Baseline     \\
        Solvable Adv      & 33.27\%  $\pm$ 1.34\%           & 70.10\%  $\pm$ 1.54\%           & 0.601 $\pm$ 0.017          & 45.39 $\pm$ 1.57          & 894,846          & 619,386    & 5.8\%       \\
        \midrule
        \multicolumn{4}{l}{\textit{Comparison of Samplers}} & & & \\
        Static        & 35.50\% $\pm$ 4.37\%           & 66.21\% $\pm$ 3.69\%           & 0.639 $\pm$ 0.030         & 42.70 $\pm$ 2.58          & 949,956    & $>$ Max Steps & 0.0\%    \\
        History       & 34.97\% $\pm$ 1.85\%           & 69.22\% $\pm$ 1.31\%           & 0.609 $\pm$ 0.027          & 44.42 $\pm$ 1.30          &  784,701    &  619,426 & 17.4\%   \\
        \midrule
        \textbf{Ours} &  \textbf{31.37\%  $\pm$ 3.51\% } & \textbf{71.66\%  $\pm$ 1.93\%} & \textbf{0.588 $\pm$ 0.026} & \textbf{46.24 $\pm$ 1.91} & \textbf{656,094} & \textbf{582,638} & \textbf{30.9\%}\\
        \bottomrule
    \end{tabular}
\end{table}

\section{Conclusion}\label{sec:conclusion}
In this paper, we propose the AlignADV framework to address the critical limitations of attack-oriented adversarial training, specifically the generation of physically unsolvable scenarios and the capability mismatch between scenario difficulty and policy competence. To improve the resolvability of generated scenarios, we employ DPO to fine-tune the pretrained trajectory generation model. Furthermore, to efficiently match the difficulty of sampled scenarios with the dynamically evolving competence of the ego driving policy, we introduce the mechanism of a behavioral fingerprint and construct a multimodal capability prediction model. The prediction model evaluates the success probability of the current policy across diverse scenarios prior to actual simulation, thereby guiding a capability-aligned dynamic curriculum sampling distribution. Comprehensive closed-loop reinforcement learning experiments demonstrate that our proposed framework reduces the required training steps by 40.6\% compared to unconstrained adversarial training baselines. Ultimately, the driving policy trained under the AlignADV framework exhibits substantial reductions in crash rates alongside persistent enhancements in route completion rates across both normal and adversarial traffic conditions.

Our findings suggest that the efficacy of adversarial training is not merely a function of attack intensity, but rather of how effectively that attack value is converted into algorithmic learnability. Despite the promising results achieved by the proposed framework, several directions are worth investigating in future research. First, to construct an even more sophisticated preference dataset for scenario generation, the rule-based privileged expert policy can be upgraded to an ensemble of advanced intelligent expert algorithms. Second, the continuous fine-tuning of the capability predictor can be integrated into the online closed-loop driving policy training paradigm. This integration will enable the prediction model to adaptively co-evolve with the autonomous driving agent, maintaining highly accurate and robust capability estimations throughout the entirety of the dynamic curriculum learning process.

\section*{Acknowledgments}
This research was supported by the National Natural Science Foundation of China (52125208, 52232015, and 524B2164). 


\appendix
\section{Rule-based expert policy} \label{sec:Appendix-expertpolicy}
To evaluate the generated scenarios, the deterministic expert policy $\pi_{expert}$ operates utilizing exact environmental ground truth. 
For dynamic interactions, the policy extracts the precise future states of all $U$ background vehicles from the simulation over a discrete prediction horizon $H$, denoted as $\mathbf{s}_{u, t+h}$ for the $u$th vehicle at future step $h$. Concurrently, the policy simulates its own future states $\hat{\mathbf{s}}_{t+h}$ via a kinematic bicycle model. 
To ensure the ego vehicle harmonizes with the overall traffic flow, the policy calibrates a dynamic reference speed $v_{dyn}$ by computing the 75th percentile of the velocity set $\mathcal{V}_{flow}$ collected from valid surrounding vehicles within a specified perception radius. 
This target speed is also strictly constrained by a predefined speed limit when the ego vehicle approaches an intersection, which is geometrically identified by evaluating the vector dot product of consecutive waypoints along the reference path to detect high curvature segments. 
Furthermore, a traffic light compliance mechanism overrides $v_{target}$ to zero immediately upon detecting a red light signal within the forward trajectory.

Based on these forecasted trajectories, the policy implements a proactive collision warning system by conducting geometric intersection tests between the oriented bounding boxes of the ego vehicle and all background entities at each forecasted step. Upon detecting an imminent spatial intersection, the policy classifies the collision threat based on the relative position and the longitudinal relative velocity.
For any predicted frontal or lateral collision threats, the policy triggers an emergency braking protocol by forcefully setting $v_{target}$ to zero.
Conversely, if a rear collision threat is identified where a background vehicle approaches rapidly from behind, and a subsequent check confirms that the forward path is completely clear, the policy executes an evasive acceleration maneuver. 

The final control execution is achieved through decoupled longitudinal and lateral proportional integral derivative controllers that track the dynamically determined targets. 
The longitudinal controller calculates the acceleration command $a_{a,t}$ to minimize the error between the final safe target speed $v_{target}$ and the current speed $v_t$. 
The lateral controller determines the steering angle $a_{\delta,t}$ by minimizing the heading error relative to a look-ahead waypoint. This waypoint is dynamically sampled from the global reference lane at a look-ahead distance proportional to the current speed $v_t$. The synthesized action $a_t = [a_{a,t}, a_{\delta,t}]^T$ is then applied to the ego vehicle.

\footnotesize
\bibliographystyle{elsarticle-harv}
\bibliography{main}

\end{document}